\documentclass[10pt,twocolumn,letterpaper]{article}

\usepackage{cvpr}
\usepackage{times}
\usepackage{epsfig}
\usepackage{graphicx}
\usepackage{amsmath}
\usepackage{amssymb}
\usepackage{booktabs}

\usepackage[pagebackref=true,breaklinks=true,letterpaper=true,colorlinks,bookmarks=false]{hyperref}

\cvprfinalcopy % *** Uncomment this line for the final submission

\def\cvprPaperID{3022} % *** Enter the CVPR Paper ID here

\usepackage{algorithm}
\usepackage{algpseudocode}
\usepackage{rotating}
\usepackage[font={small}]{caption}
\usepackage{subcaption}
\usepackage{enumitem}
\usepackage{afterpage}
\usepackage{multirow}
\usepackage{epigraph}
\usepackage{pgfplots}

\newcommand{\norm}[1]{\left\lVert #1 \right\rVert}

\usepackage{capt-of}

\newcommand{\softmax}{\text{Softmax}}

\newcommand{\N}{\mathcal{N}}
\newcommand{\cL}{\mathcal{L}}

\newcommand{\f}{\mathbf{f}}
\newcommand{\x}{\mathbf{x}}

\newcommand{\y}{\mathbf{y}}

\newcommand{\W}{\mathbf{W}}

\newcommand{\fig}[1]{Figure~\ref{fig:#1}}
\newcommand{\tbl}[1]{Table~\ref{tbl:#1}}
\newcommand{\sct}[1]{Section~\ref{sec:#1}}
\newcommand{\eqn}[1]{(\ref{eqn:#1})}

\graphicspath{{figures/}}

\begin{document}

\title{Multi-Task Learning Using Uncertainty to Weigh Losses\\for Scene Geometry and Semantics}

% The \author macro works with any number of authors. There are two
% commands used to separate the names and addresses of multiple
% authors: \And and \AND.
%
% Using \And between authors leaves it to LaTeX to determine where to
% break the lines. Using \AND forces a line break at that point. So,
% if LaTeX puts 3 of 4 authors names on the first line, and the last
% on the second line, try using \AND instead of \And before the third
% author name.

\author{
  Alex Kendall\\
  University of Cambridge\\
  \texttt{agk34@cam.ac.uk} \\
  \and
  Yarin Gal\\
  University of Oxford\\
  \texttt{yarin@cs.ox.ac.uk} \\
  \and
  Roberto Cipolla\\
  University of Cambridge\\
  \texttt{rc10001@cam.ac.uk} \\
}

\maketitle

%%%%%%%%% ABSTRACT
\begin{abstract}

Numerous deep learning applications benefit from multi-task learning with multiple regression and classification objectives. In this paper we make the observation that the performance of such systems is strongly dependent on the relative weighting between each task's loss. Tuning these weights by hand is a difficult and expensive process, making multi-task learning prohibitive in practice. We propose a principled approach to multi-task deep learning which weighs multiple loss functions by considering the homoscedastic uncertainty of each task. This allows us to simultaneously learn various quantities with different units or scales in both classification and regression settings. We demonstrate our model learning per-pixel depth regression, semantic and instance segmentation from a monocular input image. Perhaps surprisingly, we show our model can learn multi-task weightings and outperform separate models trained individually on each task.

\end{abstract}

%%%%%%%%% BODY TEXT
\section{Introduction}
\label{sec:intro}

\begin{figure*}[t]
\begin{center}
		\includegraphics[width=\linewidth]{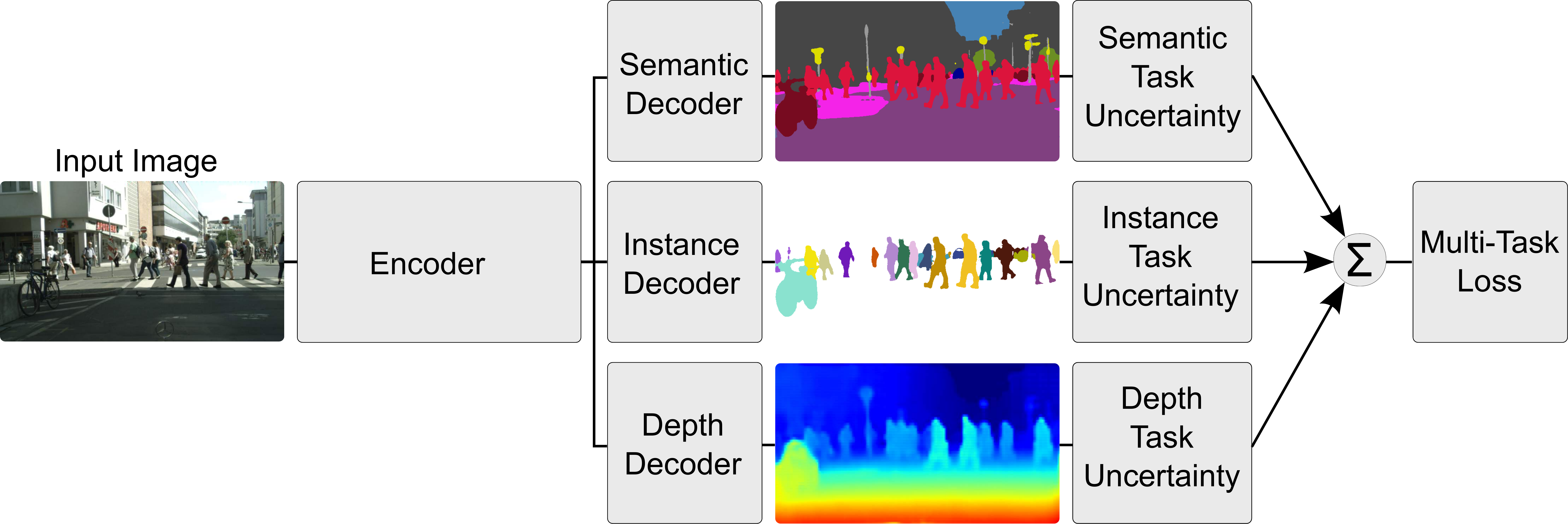}    
\end{center}
\vspace{-3mm}
   \caption{\textbf{Multi-task deep learning.} We derive a principled way of combining multiple regression and classification loss functions for multi-task learning. Our architecture takes a single monocular RGB image as input and produces a pixel-wise classification, an instance semantic segmentation and an estimate of per pixel depth. Multi-task learning can improve accuracy over separately trained models because cues from one task, such as depth, are used to regularize and improve the generalization of another domain, such as segmentation.}
\label{fig:teaser}
\vspace{-2mm}
\end{figure*}

Multi-task learning aims to improve learning efficiency and prediction accuracy by learning multiple objectives from a shared representation \cite{caruana1998multitask}. Multi-task learning is prevalent in many applications of machine learning -- from computer vision \cite{kokkinos2016ubernet} to natural language processing \cite{collobert2008unified} to speech recognition \cite{huang2013cross}.

We explore multi-task learning within the setting of visual scene understanding in computer vision. Scene understanding algorithms must understand both the geometry and semantics of the scene at the same time. This forms an interesting multi-task learning problem because scene understanding involves joint learning of various regression and classification tasks with different units and scales. Multi-task learning of visual scene understanding is of crucial importance in systems where long computation run-time is prohibitive, such as the ones used in robotics. Combining all tasks into a single model reduces computation and allows these systems to run in real-time.

Prior approaches to simultaneously learning multiple tasks use a na{\"i}ve weighted sum of losses, where the loss weights are uniform, or manually tuned \cite{sermanet2013overfeat,kokkinos2016ubernet,eigen2015predicting}. However, we show that performance is highly dependent on an appropriate choice of weighting between each task's loss. Searching for an optimal weighting is prohibitively expensive and difficult to resolve with manual tuning. We observe that the optimal weighting of each task is dependent on the measurement scale (e.g. meters, centimetres or millimetres) and ultimately the magnitude of the task's noise. 

In this work we propose a principled way of combining multiple loss functions to simultaneously learn multiple objectives using homoscedastic uncertainty. We interpret homoscedastic uncertainty as task-dependent weighting and show how to derive a principled multi-task loss function which can learn to balance various regression and classification losses. Our method can learn to balance these weightings optimally, resulting in superior performance, compared with learning each task individually. 

Specifically, we demonstrate our method in learning scene geometry and semantics with three tasks. Firstly, we learn to classify objects at a pixel level, also known as semantic segmentation \cite{long2014fully,badrinarayanan2015segnet,YuKoltun2016,chen2014semantic,zheng2015conditional}. Secondly, our model performs instance segmentation, which is the harder task of segmenting separate masks for each individual object in an image (for example, a separate, precise mask for each individual car on the road) \cite{pinheiro2015learning,hariharan2014hypercolumns,dai2016instance,bai2016deep}. This is a more difficult task than semantic segmentation, as it requires not only an estimate of each pixel's class, but also which object that pixel belongs to. It is also more complicated than object detection, which often predicts object bounding boxes alone \cite{girshick2014rich}. Finally, our model predicts pixel-wise metric depth. Depth by recognition has been demonstrated using dense prediction networks with supervised \cite{eigen2015predicting} and unsupervised \cite{garg2016unsupervised} deep learning. However it is very hard to estimate depth in a way which generalises well. We show that we can improve our estimation of geometry \textit{and} depth by using semantic labels and multi-task deep learning.

In existing literature, separate deep learning models would be used to learn depth regression, semantic segmentation and instance segmentation to create a complete scene understanding system. Given a single monocular input image, our system is the first to produce a semantic segmentation, a dense estimate of metric depth and an instance level segmentation jointly (\fig{teaser}). While other vision models have demonstrated multi-task learning, we show how to learn to combine semantics and geometry. Combining these tasks into a single model ensures that the model agrees between the separate task outputs while reducing computation. Finally, we show that using a shared representation with multi-task learning improves performance on various metrics, making the models more effective.

In summary, the key contributions of this paper are:
\begin{enumerate}[topsep=0pt,itemsep=0.5ex,partopsep=0ex,parsep=0ex]
\item a novel and principled multi-task loss to simultaneously learn various classification and regression losses of varying quantities and units using homoscedastic task uncertainty,
\item a unified architecture for semantic segmentation, instance segmentation and depth regression,
\item demonstrating the importance of loss weighting in multi-task deep learning and how to obtain superior performance compared to equivalent separately trained models.
\end{enumerate}

%We benchmark our method on the challenging CityScapes dataset for road scene understanding in section \ref{sec:benchmark}. We set a new instance segmentation benchmark on this dataset, demonstrating the efficacy of our approach.

\begin{figure*}[t]
\centering
\begin{subfigure}[c]{0.5\linewidth}
\resizebox{\linewidth}{!}{
\begin{tikzpicture}
\pgfplotsset{
    compat=1.3,
    scale only axis,
    height=5cm,
    width=10cm, % \textwidth minus width of longest label text minus label offset
    legend style={at={(0.03,0.15)},anchor=west},
}

\pgfplotsset{
  axis line style={red},
  every axis label/.append style ={red},
  every tick label/.append style={red}  
}

\begin{axis}[
  axis y line*=left,
  axis x line*=top,
  ymin=45, ymax=62,
  x dir=reverse,
  xlabel=Classification Weight,
  ylabel=IoU Classification ($\%$),
  xmin=0, xmax=1,
]
\addplot[mark=x,red,very thick]
  coordinates{
    (1.0,59.4)
    (0.975,59.5)
    (0.95,59.9)
    (0.9,60.1)
    (0.85,60.4)
    (0.8,59.6)
    (0.7,59.0)
    (0.5,56.3)
    (0.2,47.2)
    (0.1,42.7)
}; \label{plot_one2}
\addlegendentry{Classification}
\end{axis}

\pgfplotsset{
  axis line style={blue},
  every axis label/.append style ={blue},
  every tick label/.append style={blue}  
}

\begin{axis}[
  axis y line*=right,
  axis x line*=bottom,
  xlabel=Depth Weight,
  xmin=0, xmax=1,
  ymin=0.566, ymax=0.65,
  ylabel=RMS Inverse Depth Error ($m^{-1}$),
  y dir=reverse,
]
\addlegendimage{/pgfplots/refstyle=plot_one2}\addlegendentry{Classification}
\addplot[mark=*,blue,very thick]
  coordinates{
    (0.025,0.664)
    (0.05,0.603)
    (0.1,0.586)
    (0.15,0.582)
    (0.2,0.577)
    (0.3,0.573)
    (0.5,0.602)
    (0.8,0.625)
    (0.9,0.628)
    (1.0,0.640)
}; \label{plot_two2}
\addlegendentry{Depth Regression}
\end{axis}
\end{tikzpicture}}
\end{subfigure}
\qquad\qquad
\begin{subfigure}[c]{0.3\linewidth}
\resizebox{\linewidth}{!}{
\begin{tabular}{cc|cc}
    \toprule
\multicolumn{2}{c|}{Task Weights} & Class & Depth \\
Class & Depth & IoU {[}$\%${]} & Err. {[}$px${]}\\
    \midrule
1.0 & 0.0 & 59.4 & - \\ %
0.975&0.025& 59.5 & 0.664 \\ % 
0.95& 0.05& 59.9 & 0.603 \\ %
0.9 & 0.1 & 60.1 & 0.586 \\ % 
0.85& 0.15& 60.4 & 0.582 \\ % 
0.8 & 0.2 & 59.6  & 0.577 \\ %
0.7 & 0.3 & 59.0 & 0.573 \\ %
0.5 & 0.5 & 56.3 & 0.602 \\ % 
0.2 & 0.8 & 47.2 & 0.625 \\ % 
0.1 & 0.9 & 42.7 & 0.628 \\ %
0.0 & 1.0 & - & 0.640 \\ %
    \midrule
\multicolumn{2}{c|}{\textbf{Learned weights}} & \multirow{3}{*}{\textbf{62.7}} & \multirow{3}{*}{\textbf{0.533}} \\
\multicolumn{2}{c|}{\textbf{with task uncertainty}} && \\
\multicolumn{2}{c|}{\textbf{(this work, \sct{mt_loss})}} && \\
    \bottomrule
\end{tabular}}
\end{subfigure}
\vspace{2pt}

(a) Comparing loss weightings when learning \textbf{semantic classification and depth regression}
%%%%%%%%%%%%%%%%%%%%%%%%%%%%%%%%%%%%%%%%%%%%%%%%%%%%%%%%%%%%%
%\noindent\rule{14cm}{0.4pt}

\begin{subfigure}[c]{0.5\linewidth}
\resizebox{\linewidth}{!}{
\begin{tikzpicture}
\pgfplotsset{
    compat=1.3,
    scale only axis,
    height=5cm,
    width=10cm, % \textwidth minus width of longest label text minus label offset
    legend style={at={(0.03,0.85)},anchor=west},
}

\pgfplotsset{
  axis line style={red},
  every axis label/.append style ={red},
  every tick label/.append style={red}  
}

\begin{axis}[
  axis y line*=left,
  axis x line*=top,
  ymin=3.7, ymax=5.0,
  x dir=reverse,
  xlabel=Instance Weight,
  ylabel=RMS Instance (px),
  xmin=0, xmax=1,
  y dir=reverse,
]
\addplot[mark=x,red,very thick]
  coordinates{
    (1.0,4.61)
    (0.75,4.52)
    (0.5,4.30)
    (0.4,4.14)
    (0.3,4.04)
    (0.2,3.83)
    (0.1,3.91)
    (0.05,4.27)
    (0.025,4.31)
}; \label{plot_one}
\addlegendentry{Instance Regression}
\end{axis}

\pgfplotsset{
  axis line style={blue},
  every axis label/.append style ={blue},
  every tick label/.append style={blue}  
}

\begin{axis}[
  axis y line*=right,
  axis x line*=bottom,
  xlabel=Depth Weight,
  xmin=0, xmax=1,
  ymin=0.59, ymax=0.7,
  ylabel=RMS Inverse Depth Error ($m^{-1}$),
  y dir=reverse,
]
\addlegendimage{/pgfplots/refstyle=plot_one}\addlegendentry{Instance Regression}
\addplot[mark=*,blue,very thick]
  coordinates{
    (0.25,0.692)
    (0.5,0.655)
    (0.6,0.641)
    (0.7,0.615)
    (0.8,0.607)
    (0.9,0.600)
    (0.95,0.607)
    (0.975,0.624)
    (1.0,0.640)
}; \label{plot_two}
\addlegendentry{Depth Regression}
\end{axis}
\end{tikzpicture}}
\end{subfigure}
\qquad\qquad
\begin{subfigure}[c]{0.3\linewidth}
\resizebox{\linewidth}{!}{
\begin{tabular}{cc|cc}
    \toprule
\multicolumn{2}{c|}{Task Weights} & Instance & Depth \\
Instance & Depth & Err. {[}$px${]} & Err. {[}$px${]}\\ 
    \midrule
1.0 & 0.0 & 4.61 &  \\
0.75 & 0.25 & 4.52 & 0.692 \\
0.5 & 0.5 & 4.30 & 0.655 \\
0.4 & 0.6 & 4.14 & 0.641 \\
0.3 & 0.7 & 4.04 & 0.615 \\
0.2 & 0.8 & 3.83 & 0.607 \\
0.1 & 0.9 & 3.91 & 0.600 \\
0.05 & 0.95 & 4.27 & 0.607 \\
0.025 & 0.975 & 4.31 & 0.624 \\
0.0 & 1.0 &  & 0.640 \\
    \midrule
\multicolumn{2}{c|}{\textbf{Learned weights}} & \multirow{3}{*}{\textbf{3.54}} & \multirow{3}{*}{\textbf{0.539}} \\
\multicolumn{2}{c|}{\textbf{with task uncertainty}} && \\
\multicolumn{2}{c|}{\textbf{(this work, \sct{mt_loss})}} && \\
    \bottomrule
\end{tabular}}
\end{subfigure}
%%%%%%%%%%%%%%%%%%%%%%%%%%%%%%%%%%%%%%%%%%%%%%%%%%%%%%%%%%%%%

(b) Comparing loss weightings when learning \textbf{instance regression and depth regression}
%\noindent\rule{14cm}{0.4pt}

   \caption{\textbf{Learning multiple tasks improves the model's representation and individual task performance}. These figures and tables illustrate the advantages of multi-task learning for (a) semantic classification and depth regression and (b) instance and depth regression. Performance of the model in individual tasks is seen at both edges of the plot where $w=0$ and $w=1$. For some balance of weightings between each task, we observe improved performance for both tasks. All models were trained with a learning rate of $0.01$ with the respective weightings applied to the losses using the loss function in \eqn{basic_loss}. Results are shown using the Tiny CityScapes validation dataset using a down-sampled resolution of $128\times256$.}
\label{fig:scale_factor}
\end{figure*}

\section{Related Work}

Multi-task learning aims to improve learning efficiency and prediction accuracy for each task, when compared to training a separate model for each task \cite{thrun1996learning,baxter2000model}. It can be considered an approach to inductive knowledge transfer which improves generalisation by sharing the domain information between complimentary tasks. It does this by using a shared representation to learn multiple tasks -- what is learned from one task can help learn other tasks \cite{caruana1998multitask}.

Fine-tuning \cite{agrawal2015learning,oquab2014learning} is a basic example of multi-task learning, where we can leverage different learning tasks by considering them as a pre-training step.
Other models alternate learning between each training task, for example in natural language processing \cite{collobert2008unified}.
Multi-task learning can also be used in a data streaming setting \cite{thrun1996learning}, or to prevent forgetting previously learned tasks in reinforcement learning \cite{kirkpatrick2017overcoming}. It can also be used to learn unsupervised features from various data sources with an auto-encoder \cite{ngiam2011multimodal}.

In computer vision there are many examples of methods for multi-task learning. Many focus on semantic tasks, such as classification and semantic segmentation \cite{liao2016understand} or classification and detection \cite{sermanet2013overfeat}. MultiNet \cite{teichmann2016multinet} proposes an architecture for detection, classification and semantic segmentation. CrossStitch networks \cite{misra2016cross} explore methods to combine multi-task neural activations. Uhrig et al. \cite{uhrig2016pixel} learn semantic and instance segmentations under a classification setting. Multi-task deep learning has also been used for geometry and regression tasks. \cite{eigen2015predicting} show how to learn semantic segmentation, depth and surface normals. PoseNet \cite{kendall2015convolutional} is a model which learns camera position and orientation. UberNet \cite{kokkinos2016ubernet} learns a number of different regression and classification tasks under a single architecture. In this work we are the first to propose a method for jointly learning depth regression, semantic and instance segmentation. Like the model of \cite{eigen2015predicting}, our model learns both semantic and geometry representations, which is important for scene understanding. However, our model learns the much harder task of instance segmentation which requires knowledge of both semantics and geometry. This is because our model must determine the class and spatial relationship for each pixel in each object for instance segmentation.

More importantly, all previous methods which learn multiple tasks simultaneously use a na{\"i}ve weighted sum of losses, where the loss weights are uniform, or crudely and manually tuned. In this work we propose a principled way of combining multiple loss functions to simultaneously learn multiple objectives using homoscedastic task uncertainty. We illustrate the importance of appropriately weighting each task in deep learning to achieve good performance and show that our method can learn to balance these weightings optimally.

\section{Multi Task Learning with Homoscedastic Uncertainty}
\label{sec:multitask}

Multi-task learning concerns the problem of optimising a model with respect to multiple objectives. It is prevalent in many deep learning problems. The naive approach to combining multi objective losses would be to simply perform a weighted linear sum of the losses for each individual task:
\begin{equation}
\label{eqn:basic_loss}
L_{total}= \sum_i w_i L_{i}.
\end{equation}
This is the dominant approach used by prior work \cite{teichmann2016multinet,sermanet2013overfeat,liao2016understand,uhrig2016pixel}, for example for dense prediction tasks \cite{kokkinos2016ubernet}, for scene understanding tasks \cite{eigen2015predicting} and for rotation (in quaternions) and translation (in meters) for camera pose \cite{kendall2015convolutional}. However, there are a number of issues with this method. Namely, model performance is extremely sensitive to weight selection, $w_i$, as illustrated in \fig{scale_factor}. These weight hyper-parameters are expensive to tune, often taking many days for each trial. Therefore, it is desirable to find a more convenient approach which is able to learn the optimal weights.

More concretely, let us consider a network which learns to predict pixel-wise depth and semantic class from an input image. In \fig{scale_factor} the two boundaries of each plot show models trained on individual tasks, with the curves showing performance for varying weights $w_i$ for each task. We observe that at some optimal weighting, the joint network performs better than separate networks trained on each task individually (performance of the model in individual tasks is seen at both edges of the plot: $w=0$ and $w=1$). At near-by values to the optimal weight the network performs worse on one of the tasks. However, searching for these optimal weightings is expensive and increasingly difficult with large models with numerous tasks. \fig{scale_factor} also shows a similar result for two regression tasks; instance segmentation and depth regression. We next show how to learn optimal task weightings using ideas from probabilistic modelling.

\subsection{Homoscedastic uncertainty as task-dependent uncertainty}
\label{sec:uncertainty}

In Bayesian modelling, there are two main types of uncertainty one can model \cite{kendall2017uncertainties}.
\begin{itemize}
\item \textit{Epistemic uncertainty} is uncertainty in the model, which captures what our model does not know due to lack of training data. It can be explained away with increased training data.
\item \textit{Aleatoric uncertainty} captures our uncertainty with respect to information which our data cannot explain. Aleatoric uncertainty can be explained away with the ability to observe all explanatory variables with increasing precision.
\end{itemize}
Aleatoric uncertainty can again be divided into two sub-categories.
\begin{itemize}
\item \textit{Data-dependent} or \textit{Heteroscedastic} uncertainty is aleatoric uncertainty which depends on the input data and is predicted as a model output.
\item \textit{Task-dependent} or \textit{Homoscedastic} uncertainty is  aleatoric uncertainty which is not dependent on the input data. It is not a model output, rather it is a quantity which stays constant for all input data and varies between different tasks. It can therefore be described as task-dependent uncertainty.
\end{itemize}
In a multi-task setting, we show that the task uncertainty captures the relative confidence between tasks, reflecting the uncertainty inherent to the regression or classification task. It will also depend on the task's representation or unit of measure.  We propose that we can use homoscedastic uncertainty as a basis for weighting losses in a multi-task learning problem.

\subsection{Multi-task likelihoods}
\label{sec:mt_loss}

In this section we derive a multi-task loss function based on maximising the Gaussian likelihood with homoscedastic uncertainty.
Let $\f^\W(\x)$ be the output of a neural network with weights $\W$ on input $\x$. We define the following probabilistic model. 
For regression tasks we define our likelihood as a Gaussian with mean given by the model output:
\begin{equation}
p(\y | \f^\W(\x)) = \N(\f^\W(\x), \sigma^2)
\end{equation}
with an observation noise scalar $\sigma$. 
For classification we often squash the model output through a softmax function, and sample from the resulting probability vector:
\begin{equation}
p(\y | \f^\W(\x)) = \softmax(\f^\W(\x)).
\end{equation}

In the case of multiple model outputs, we often define the likelihood to factorise over the outputs, given some sufficient statistics. We define $\f^\W(\x)$ as our sufficient statistics, and obtain the following multi-task likelihood:
\begin{equation}
p(\y_1, ..., \y_K | \f^\W(\x)) = p(\y_1 | \f^\W(\x)) ... p(\y_K | \f^\W(\x))
\end{equation}
with model outputs $\y_1, ..., \y_K$ (such as semantic segmentation, depth regression, etc).

In \textit{maximum likelihood} inference, we maximise the log likelihood of the model.
In regression, for example, the log likelihood can be written as
\begin{equation}
\log p(\y | \f^{\W}(\x)) \propto 
-\frac{1}{2 \sigma^2}||\y - \f^{\W}(\x)||^2 - \log \sigma
\end{equation}
for a Gaussian likelihood (or similarly for a Laplace likelihood)
% \begin{align*}
% \log p(\y | \f^{\W}(\x)) \propto 
% -\frac{1}{\sigma^2}||\y - \f^{\W}(\x)|| - \log \sigma^2 
% \end{align*}
% for a Laplace likelihood,
with $\sigma$ the model's observation noise parameter -- capturing how much noise we have in the outputs. We then maximise the log likelihood with respect to the model parameters $\W$ and observation noise parameter $\sigma$.

Let us now assume that our model output is composed of two vectors $\y_1$ and $\y_2$, each following a Gaussian distribution:
\begin{equation}
\begin{split}
p(\y_1, \y_2 | \f^\W(\x)) &= 
p(\y_1 | \f^\W(\x)) \cdot p(\y_2 | \f^\W(\x)) \\
&= 
\N(\y_1; \f^\W(\x), \sigma_1^2) \cdot
\N(\y_2; \f^\W(\x), \sigma_2^2).
\end{split}
\end{equation}

This leads to the \textit{minimisation} objective, $\cL(\W, \sigma_1, \sigma_2)$, (our loss) for our multi-output model:
\begin{equation}
\begin{split}
 &= 
- \log p(\y_1, \y_2 | \f^\W(\x)) \\
&\propto
\frac{1}{2 \sigma_1^2} ||\y_1 - \f^{\W}(\x)||^2
+ \frac{1}{2 \sigma_2^2} ||\y_2 - \f^{\W}(\x)||^2 
+ \log \sigma_1 \sigma_2 \\
&= \frac{1}{2 \sigma_1^2} \cL_1(\W) 
+ \frac{1}{2 \sigma_2^2} \cL_2(\W)
+ \log \sigma_1 \sigma_2
\end{split}
\end{equation}
Where we wrote $\cL_1(\W) = ||\y_1 - \f^{\W}(\x)||^2$ for the loss of the first output variable, and similarly for $\cL_2(\W)$.

% Rewriting this objective 

We interpret minimising this last objective with respect to $\sigma_1$ and $\sigma_2$ as learning the relative weight of the losses $\cL_1(\W)$ and $\cL_2(\W)$ adaptively, based on the data. As $\sigma_1$ -- the noise parameter for the variable $\y_1$ -- increases, we have that the weight of $\cL_1(\W)$ decreases. On the other hand, as the noise decreases, we have that the weight of the respective objective increases. The noise is discouraged from increasing too much (effectively ignoring the data) by the last term in the objective, which acts as a regulariser for the noise terms.

This construction can be trivially extended to multiple regression outputs. However, the extension to classification likelihoods is more interesting. We adapt the classification likelihood to squash a \textit{scaled} version of the model output through a softmax function:
\begin{equation}
p(\y | \f^\W(\x), \sigma) = \softmax( \frac{1}{\sigma^2} \f^\W(\x))
\end{equation}
with a positive scalar $\sigma$. 
This can be interpreted as a Boltzmann distribution (also called Gibbs distribution) where the input is scaled by $\sigma^2$ (often referred to as \textit{temperature}). This scalar is either fixed or can be learnt, where the parameter's magnitude determines how `uniform' (flat) the discrete distribution is. This relates to its uncertainty, as measured in entropy.
The log likelihood for this output can then be written as 
\begin{equation}
\begin{split}
\log p(\y = c | \f^\W(\x), \sigma) &= \frac{1}{\sigma^2} f^\W_c(\x) \\
&- \log \sum_{c'} \exp \bigg( \frac{1}{\sigma^2} f^\W_{c'}(\x) \bigg)
\end{split}
\end{equation}
with $f^\W_c(\x)$ the $c$'th element of the vector $\f^\W(\x)$. 
% We can rewrite this log likelihood in terms of $\cL(\W) = - \log \softmax (\f^\W(\x))$, the cross entropy loss:
% \begin{align*}
% \log p(\y = c | \f^\W(\x), \sigma) 
% &= 
% - \frac{1}{\sigma^2} \cL(\W)
% + \frac{1}{\sigma^2} \log \sum_{c'} \exp \bigg( f^\W_{c'}(\x) \bigg)
% - \log \sum_{c'} \exp \bigg( \frac{1}{\sigma^2} f^\W_{c'}(\x) \bigg)
% \\
% % &= 
% % \frac{1}{\sigma^2} \cL(\W)
% % + \log \bigg( \sum_{c'} \exp \bigg( f^\W_{c'}(\x) \bigg) \bigg)^{\frac{1}{\sigma^2} }
% % - \log \sum_{c'} \exp \bigg( \frac{1}{\sigma^2} f^\W_{c'}(\x) \bigg)
% % \\
% &= 
% -\frac{1}{\sigma^2} \cL(\W)
% - \log \frac{
% \sum_{c'} \exp \bigg( \frac{1}{\sigma^2} f^\W_{c'}(\x) \bigg)
% }{
% \bigg( \sum_{c'} \exp \bigg( f^\W_{c'}(\x) \bigg) \bigg)^{\frac{1}{\sigma^2} }
% }.
% \end{align*}

Next, assume that a model's multiple outputs are composed of a continuous output $\y_1$ and a discrete output $\y_2$, modelled with a Gaussian likelihood and a softmax likelihood, respectively. Like before, the joint loss, $\cL(\W, \sigma_1, \sigma_2)$, is given as:
\begin{equation}
\begin{aligned}
 &= 
-\log p(\y_1, \y_2=c | \f^\W(\x)) \\
&= 
-\log \N(\y_1; \f^\W(\x), \sigma_1^2) \cdot
\softmax(\y_2=c; \f^\W(\x), \sigma_2) \\
&=
\frac{1}{2 \sigma_1^2} ||\y_1 - \f^{\W}(\x)||^2
+ \log \sigma_1
- \log p(\y_2 = c | \f^\W(\x), \sigma_2)
 \\
&= \frac{1}{2 \sigma_1^2} \cL_1(\W) 
+ \frac{1}{\sigma_2^2} \cL_2(\W)
+ \log \sigma_1 \\
&\quad\quad\quad\quad + \log \frac{
\sum_{c'} \exp \bigg( \frac{1}{\sigma_2^2} f^\W_{c'}(\x) \bigg)
}{
\bigg( \sum_{c'} \exp \bigg( f^\W_{c'}(\x) \bigg) \bigg)^{\frac{1}{\sigma_2^2} }
}
\\
&\approx \frac{1}{2 \sigma_1^2} \cL_1(\W) 
+ \frac{1}{\sigma_2^2} \cL_2(\W)
+ \log \sigma_1
+ \log \sigma_2,
\end{aligned}
\end{equation}
where again we write $\cL_1(\W) = ||\y_1 - \f^{\W}(\x)||^2$ for the Euclidean loss of $\y_1$, write $\cL_2(\W) = -\log \softmax (\y_2, \f^\W(\x))$ for the cross entropy loss of $\y_2$ (with $\f^{\W}(\x)$ not scaled), and optimise with respect to $\W$ as well as $\sigma_1$, $\sigma_2$.
In the last transition we introduced the explicit simplifying assumption $\frac{1}{\sigma_2} \sum_{c'} \exp \bigg( \frac{1}{\sigma_2^2} f^\W_{c'}(\x) \bigg) \approx \bigg( \sum_{c'} \exp \bigg( f^\W_{c'}(\x) \bigg) \bigg)^{\frac{1}{\sigma_2^2} }$ which becomes an equality when $\sigma_2 \rightarrow 1$. This has the advantage of simplifying the optimisation objective, as well as empirically improving results.

This last objective can be seen as learning the relative weights of the losses for each output. Large scale values $\sigma_2$ will decrease the contribution of $\cL_2(\W)$, whereas small scale $\sigma_2$ will increase its contribution. The scale is regulated by the last term in the equation. The objective is penalised when setting $\sigma_2$ too large. % (with the last term contributing a constant value $\log C$ -- with $C$ classes -- to the loss). 

This construction can be trivially extended to arbitrary combinations of discrete and continuous loss functions, allowing us to learn the relative weights of each loss in a principled and well-founded way. This loss is smoothly differentiable, and is well formed such that the task weights will not converge to zero. In contrast, directly learning the weights using a simple linear sum of losses \eqn{basic_loss} would result in weights which quickly converge to zero. In the following sections we introduce our experimental model and present empirical results.

In practice, we train the network to predict the log variance, $s := \log \sigma^2$.
This is because it is more numerically stable than regressing the variance, $\sigma^2$, as the loss avoids any division by zero. The exponential mapping also allows us to regress unconstrained scalar
values, where $\exp(-s)$ is resolved to the positive domain giving valid values for variance.

%Our final learning objective is,
%\begin{equation}
%\cL = 2 \exp(-s_C)  \cL_C 
%+ \exp(-s_I)  \cL_I
%+ \exp(-s_D)  \cL_D
%+ s_C + s_I + s_D,
%\end{equation}

%%%%%%%%%%%%%%%%%%%%%%%%%%%%%%%%%%%%%%%%%%%%%%%%%%%%%%%%%%%%%%%%%%%%%%%%%%%%%%%%%%%%%%%%%%%%%%

\section{Scene Understanding Model}
\label{sec:arch}

To understand semantics and geometry we first propose an architecture which can learn regression and classification outputs, at a pixel level. Our architecture is a deep convolutional encoder decoder network \cite{badrinarayanan2015segnet}. Our model consists of a number of convolutional encoders which produce a shared representation, followed by a corresponding number of task-specific convolutional decoders. A high level summary is shown in \fig{teaser}.

The purpose of the encoder is to learn a deep mapping to produce rich, contextual features, using domain knowledge from a number of related tasks. Our encoder is based on DeepLabV3 \cite{chen2017rethinking}, which is a state of the art semantic segmentation framework. We use ResNet101 \cite{he2015deep} as the base feature encoder, followed by an Atrous Spatial Pyramid Pooling (ASPP) module \cite{chen2017rethinking} to increase contextual awareness. We apply dilated convolutions in this encoder, such that the resulting feature map is sub-sampled by a factor of 8 compared to the input image dimensions. 

We then split the network into separate decoders (with separate weights) for each task. The purpose of the decoder is to learn a mapping from the shared features to an output. Each decoder consists of a $3\times3$ convolutional layer with output feature size 256, followed by a $1\times1$ layer regressing the task's output. 
%We use in-place activated batch normalisation \cite{bulo2017place} for all layers in our model. 
Further architectural details are described in Appendix \ref{apdx:arch}.

\textbf{Semantic Segmentation.}
We use the cross-entropy loss to learn pixel-wise class probabilities, averaging the loss over the pixels with semantic labels in each mini-batch.
%$\cL_{class}(\W) = \frac{1}{|N_C|} \sum_{N_C} -\log \softmax (\y_i, \f^\W(\x))$. We average the loss over the pixels with semantic labels in each mini-batch, $N_C$. 

\textbf{Instance Segmentation.}
An intuitive method for defining which instance a pixel belongs to is an association to the instance's centroid.
% In a practical setting, it is extremely unlikely two instances share the same centroid. 
We use a regression approach for instance segmentation \cite{liang2015proposal}. This approach is inspired by \cite{leibe2008robust} which identifies instances using Hough votes from object parts. In this work we extend this idea by using votes from individual pixels using deep learning. We learn an instance vector, $\hat{x}_n$, for each pixel coordinate, $c_n$, which points to the centroid of the pixel's instance, $i_n$, such that $i_n=\hat{x}_n+c_n$. We train this regression with an $L_1$ loss using ground truth labels $x_n$, averaged over all labelled pixels, $N_I$, in a mini-batch: $\cL_{Instance} = \frac{1}{|N_I|} \sum_{N_I} \norm{x_n - \hat{x}_n}_1$.

\begin{figure}[t]
\begin{center}
\begin{subfigure}[t]{0.48\linewidth}
  \includegraphics[width=\linewidth]{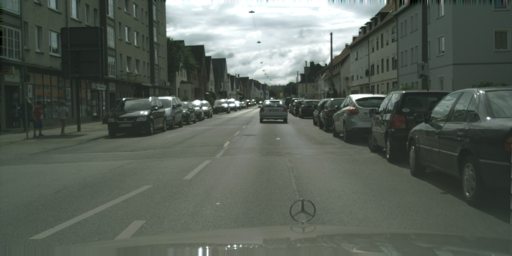}
  \caption{Input Image}
\end{subfigure}
\begin{subfigure}[t]{0.48\linewidth}
  \includegraphics[width=\linewidth]{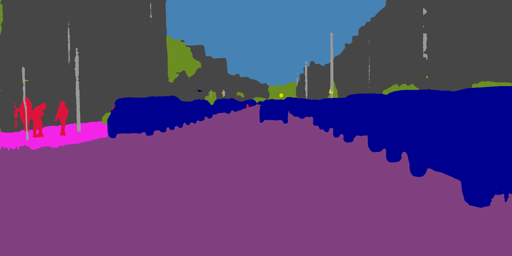}
  \caption{Semantic Segmentation}
\end{subfigure}
\begin{subfigure}[t]{0.48\linewidth}
  \includegraphics[width=\linewidth]{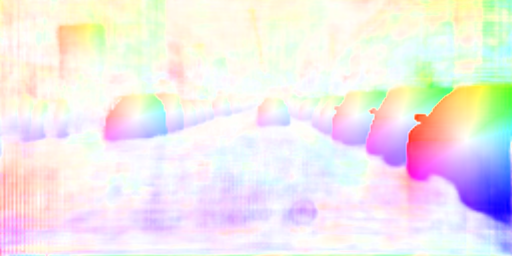}
  \caption{Instance vector regression}
\end{subfigure}
\begin{subfigure}[t]{0.48\linewidth}
  \includegraphics[width=\linewidth]{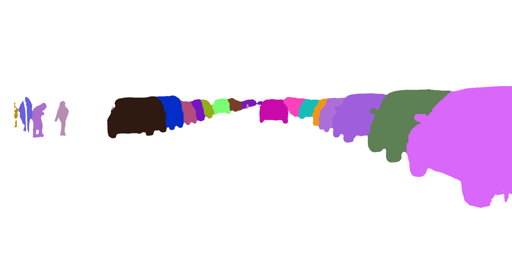}
  \caption{Instance Segmentation}
\end{subfigure}
\end{center}
   \caption{\textbf{Instance centroid regression method.} For each pixel, we regress a vector pointing to the instance's centroid. The loss is only computed over pixels which are from instances. We visualise (c) by representing colour as the orientation of the instance vector, and intensity as the magnitude of the vector.}
\label{fig:instance}
\end{figure}

\fig{instance} details the representation we use for instance segmentation. \fig{instance}(a) shows the input image and a mask of the pixels which are of an instance class (at test time inferred from the predicted semantic segmentation). \fig{instance}(b) and \fig{instance}(c) show the ground truth and predicted instance vectors for both $x$ and $y$ coordinates. We then cluster these votes using OPTICS \cite{ankerst1999optics}, resulting in the predicted instance segmentation output in \fig{instance}(d).

\begin{figure}[t]
\begin{center}
\begin{subfigure}[t]{0.49\linewidth}
  \includegraphics[width=\linewidth,trim={350px 120px 0 100px},clip]{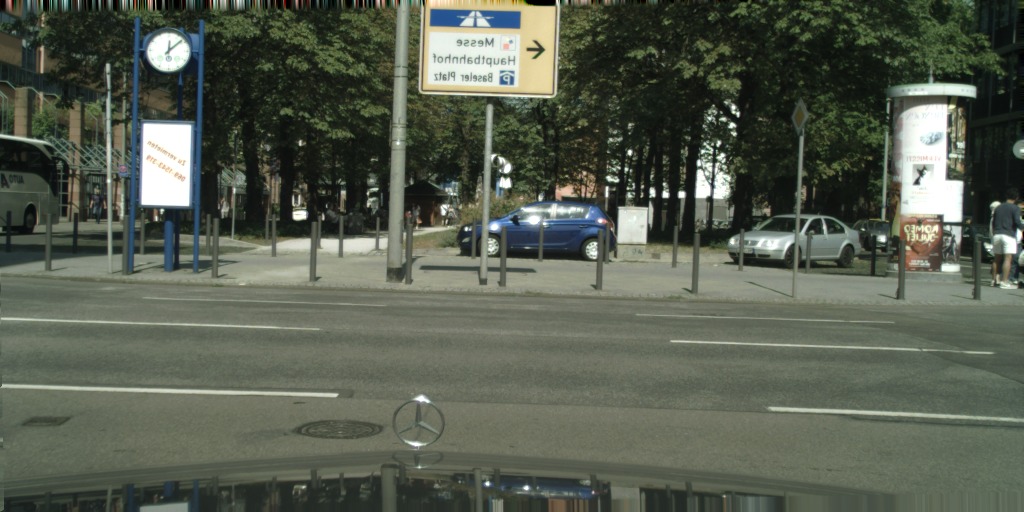}
  \caption{Input Image}
\end{subfigure}
\begin{subfigure}[t]{0.49\linewidth}
  \includegraphics[width=\linewidth,trim={350px 120px 0 100px},clip]{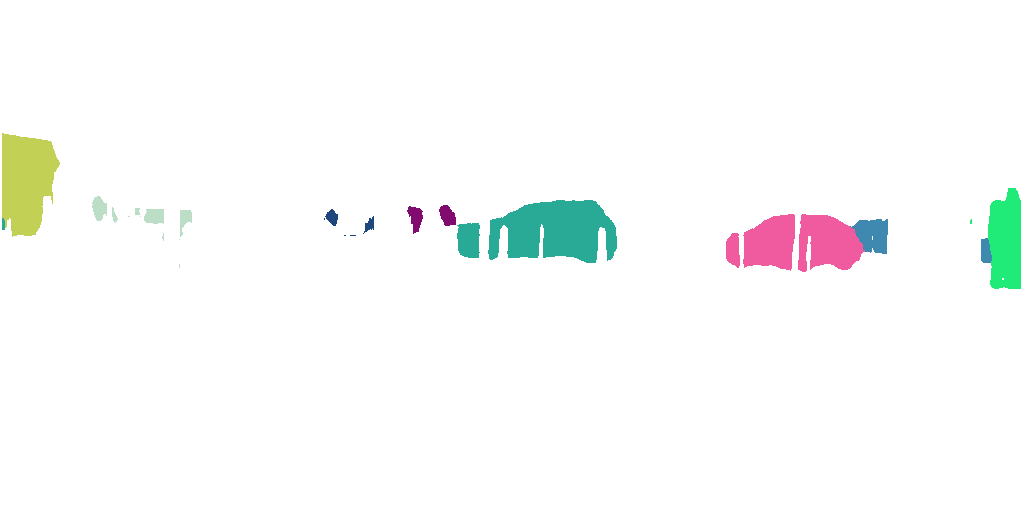}
  \caption{Instance Segmentation}
\end{subfigure}
\end{center}
   \caption{This example shows two cars which are occluded by trees and lampposts, making the instance segmentation challenging. Our instance segmentation method can handle occlusions effectively. We can correctly handle segmentation masks which are split by occlusion, yet part of the same instance, by incorporating semantics and geometry.}
\label{fig:instancemask}
\end{figure}

One of the most difficult cases for instance segmentation algorithms to handle is when the instance mask is split due to occlusion. \fig{instancemask} shows that our method can handle these situations, by allowing pixels to vote for their instance centroid with geometry. Methods which rely on watershed approaches \cite{bai2016deep}, or instance edge identification approaches fail in these scenarios.

To obtain segmentations for each instance, we now need to estimate the instance centres, $\hat{i}_n$. We propose to consider the estimated instance vectors, $\hat{x}_n$, as votes in a Hough parameter space and use a clustering algorithm to identify these instance centres. OPTICS \cite{ankerst1999optics}, is an efficient density based clustering algorithm. It is able to identify an unknown number of multi-scale clusters with varying density from a given set of samples. We chose OPICS for two reasons. Crucially, it does not assume knowledge of the number of clusters like algorithms such as k-means \cite{macqueen1967some}. Secondly, it does not assume a canonical instance size or density like discretised binning approaches \cite{comaniciu2002mean}. Using OPTICS, we cluster the points $c_n+\hat{x}_n$ into a number of estimated instances, $\hat{i}$. We can then assign each pixel, $p_n$ to the instance closest to its estimated instance vector, $c_n+\hat{x}_n$.

\textbf{Depth Regression.}
We train with supervised labels using pixel-wise metric inverse depth using a $L_1$ loss function: $\cL_{Depth} = \frac{1}{|N_D|} \sum_{N_D} \norm{d_n-\hat{d_n}}_1$. Our architecture estimates inverse depth, $\hat{d}_n$, because it can represent points at infinite distance (such as sky). We can obtain inverse depth labels, $d_n$, from a RGBD sensor or stereo imagery. Pixels which do not have an inverse depth label are ignored in the loss.

\section{Experiments}
\label{sec:experiments}

\begin{table*}[t]
\centering
%\resizebox{\textwidth}{!}{
\begin{tabular}{l|ccc|c|c|c}
    \toprule
& \multicolumn{3}{c|}{Task Weights} & Segmentation & Instance & Inverse Depth \\
Loss & Seg. & Inst. & Depth & IoU {[}$\%${]} & Mean Error {[}$px${]} & Mean Error {[}$px${]}\\ 
    \midrule
Segmentation only & 1 & 0 & 0 & 59.4\% &-&-\\
Instance only & 0 & 1 & 0 &-& 4.61 &-\\
Depth only & 0 & 0 & 1 &-&-& 0.640 \\ 
    \midrule
Unweighted sum of losses & 0.333 & 0.333 & 0.333 & 50.1\% & 3.79 & 0.592 \\
    \midrule
Approx. optimal weights & 0.89 & 0.01 & 0.1 & 62.8\% & 3.61 & 0.549 \\
    \midrule
2 task uncertainty weighting & \checkmark & \checkmark & & 61.0\% & \textbf{3.42} & - \\%done
2 task uncertainty weighting & \checkmark & & \checkmark & 62.7\% & - & 0.533 \\%done
2 task uncertainty weighting & & \checkmark & \checkmark & - & 3.54 & 0.539 \\%underway
    \midrule
3 task uncertainty weighting & \checkmark & \checkmark & \checkmark & \textbf{63.4\%} & 3.50 & \textbf{0.522} \\ %underway
    \bottomrule
\end{tabular}
\caption{Quantitative improvement when learning semantic segmentation, instance segmentation and depth with our multi-task loss. Experiments were conducted on the Tiny CityScapes dataset (sub-sampled to a resolution of $128\times 256$). Results are shown from the validation set. We observe an improvement in performance when training with our multi-task loss, over both single-task models and weighted losses. Additionally, we observe an improvement when training on all three tasks ($3\times \checkmark$) using our multi-task loss, compared with all pairs of tasks alone (denoted by $2\times \checkmark$). This shows that our loss function can automatically learn a better performing weighting between the tasks than the baselines.}
\label{tbl:multitasks}
\end{table*}

We demonstrate the efficacy of our method on CityScapes \cite{Cordts2016Cityscapes}, a large dataset for road scene understanding. It comprises of stereo imagery, from automotive grade stereo cameras with a $22cm$ baseline, labelled with instance and semantic segmentations from 20 classes. Depth images are also provided, labelled using SGM \cite{hirschmuller2008stereo}, which we treat as pseudo ground truth. Additionally, we assign zero inverse depth to pixels labelled as sky. The dataset was collected from a number of cities in fine weather and consists of 2,975 training and 500 validation images at $2048\times1024$ resolution. 1,525 images are withheld for testing on an online evaluation server.

Further training details, and optimisation hyperparameters, are provided in Appendix \ref{apdx:arch}.

\subsection{Model Analysis}
\label{sec:tasks}

In \tbl{multitasks} we compare individual models to multi-task learning models using a na{\"i}ve weighted loss or the task uncertainty weighting we propose in this paper. To reduce the computational burden, we train each model at a reduced resolution of $128\times256$ pixels, over $50,000$ iterations. When we downsample the data by a factor of four, we also need to scale the disparity labels accordingly. \tbl{multitasks} clearly illustrates the benefit of multi-task learning, which obtains significantly better performing results than individual task models. For example, using our method we improve classification results from $59.4\%$ to $63.4\%$.

We also compare to a number of na{\"i}ve multi-task losses. We compare weighting each task equally and using approximately optimal weights. Using a uniform weighting results in poor performance, in some cases not even improving on the results from the single task model. Obtaining approximately optimal weights is difficult with increasing number of tasks as it requires an expensive grid search over parameters. However, even these weights perform worse compared with our proposed method. \fig{scale_factor} shows that using task uncertainty weights can even perform better compared to optimal weights found through fine-grained grid search. We believe that this is due to two reasons. First, grid search is restricted in accuracy by the resolution of the search. Second, optimising the task weights using a homoscedastic noise term allows for the weights to be dynamic during training. In general, we observe that the uncertainty term decreases during training which improves the optimisation process.

In Appendix \ref{apdx:analysis} we find that our task-uncertainty loss is robust to the initialisation chosen for the parameters. These quickly converge to a similar optima in a few hundred training iterations. We also find the resulting task weightings varies throughout the course of training. For our final model (in \tbl{benchmarks}), at the end of training, the losses are weighted with the ratio 43 : 1 : 0.16 for semantic segmentation, depth regression and instance segmentation, respectively.

\begin{figure*}[t]
\resizebox{\linewidth}{!}{
\begin{subfigure}[t]{0.24\linewidth}
\begin{center}
		\includegraphics[width=\linewidth,trim={0px 60px 0 0px},clip]{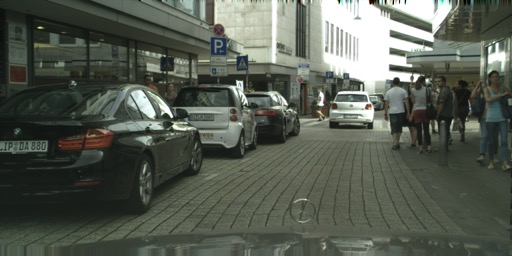}
		\includegraphics[width=\linewidth,trim={0px 60px 0 0px},clip]{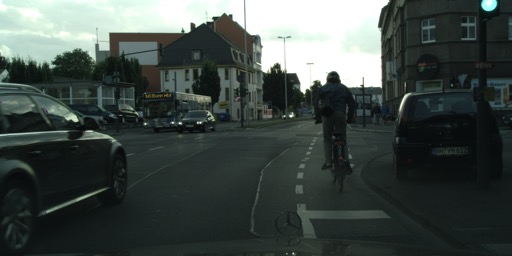}
		\includegraphics[width=\linewidth,trim={0px 60px 0 0px},clip]{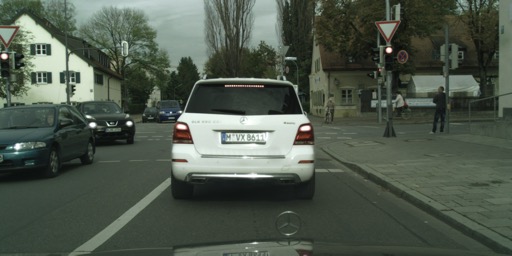}
		\includegraphics[width=\linewidth,trim={0px 60px 0 0px},clip]{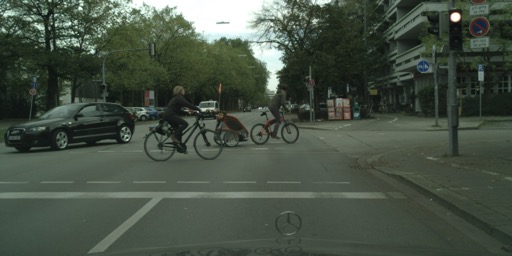}
  \caption{Input image}
\end{center}
\end{subfigure}
\begin{subfigure}[t]{0.24\linewidth}
\begin{center}
		\includegraphics[width=\linewidth,trim={0px 60px 0 0px},clip]{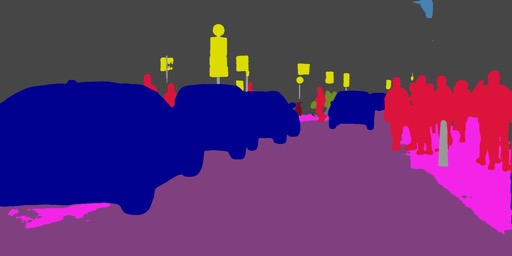}
		\includegraphics[width=\linewidth,trim={0px 60px 0 0px},clip]{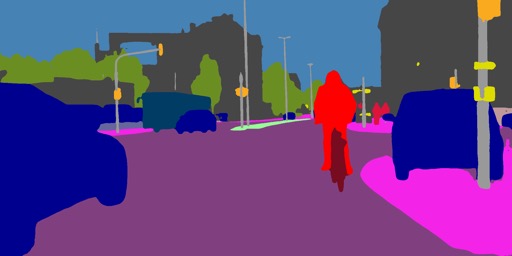}
		\includegraphics[width=\linewidth,trim={0px 60px 0 0px},clip]{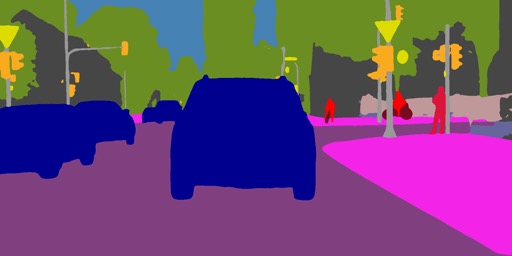}
		\includegraphics[width=\linewidth,trim={0px 60px 0 0px},clip]{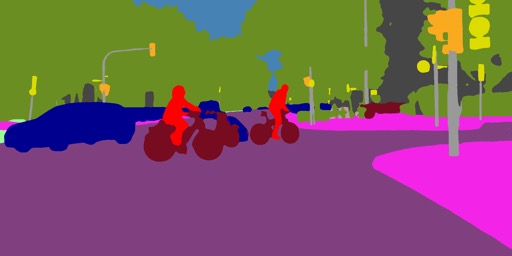}
  \caption{Segmentation output}
\end{center}
\end{subfigure}
\begin{subfigure}[t]{0.24\linewidth}
\begin{center}
		\includegraphics[width=\linewidth,trim={0px 60px 0 0px},clip]{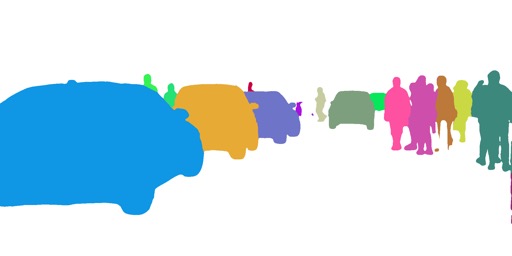}
		\includegraphics[width=\linewidth,trim={0px 60px 0 0px},clip]{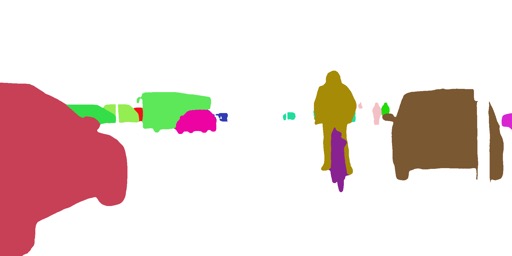}
		\includegraphics[width=\linewidth,trim={0px 60px 0 0px},clip]{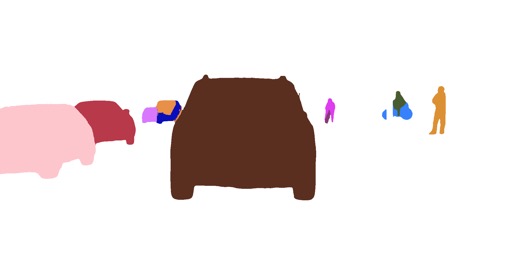}
		\includegraphics[width=\linewidth,trim={0px 60px 0 0px},clip]{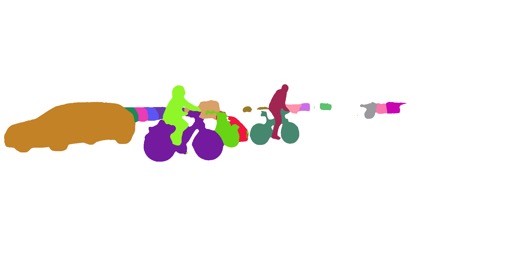}
  \caption{Instance output}
\end{center}
\end{subfigure}
\begin{subfigure}[t]{0.24\linewidth}
\begin{center}
		\includegraphics[width=\linewidth,trim={0px 60px 0 0px},clip]{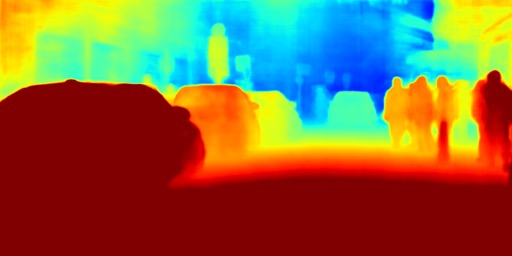}
		\includegraphics[width=\linewidth,trim={0px 60px 0 0px},clip]{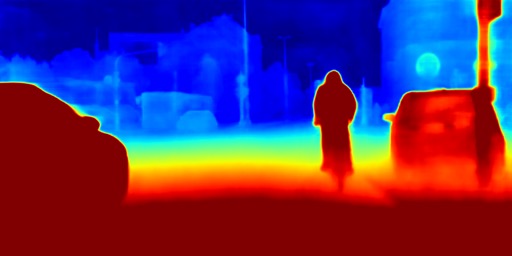}
		\includegraphics[width=\linewidth,trim={0px 60px 0 0px},clip]{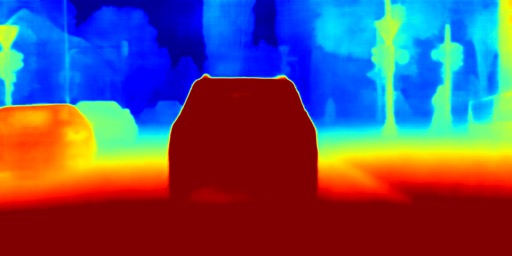}
		\includegraphics[width=\linewidth,trim={0px 60px 0 0px},clip]{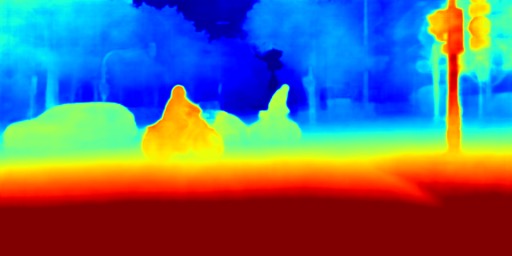}
  \caption{Depth output}
\end{center}
\end{subfigure}}
	\caption{\textbf{Qualitative results for multi-task learning of geometry and semantics for road scene understanding}. Results are shown on test images from the CityScapes dataset using our multi-task approach with a single network trained on all tasks. We observe that multi-task learning improves the smoothness and accuracy for depth perception because it learns a representation that uses cues from other tasks, such as segmentation (and vice versa).}
	\label{fig:cityscapesqual}
	\vspace{-4mm}
\end{figure*}

Finally, we benchmark our model using the full-size CityScapes dataset. In \tbl{benchmarks} we compare to a number of other state of the art methods in all three tasks. Our method is the first model which completes all three tasks with a single model. We compare favourably with other approaches, outperforming many which use comparable training data and inference tools. \fig{cityscapesqual} shows some qualitative examples of our model.

% Visualise model uncertainty...

\begin{table*}[t]
	\centering
	\resizebox{\linewidth}{!}{
		\begin{tabular}{l|c|c|c|c|c|c|c|c|c|c}
        \toprule
			& \multicolumn{4}{c|}{Semantic Segmentation} & \multicolumn{4}{c|}{Instance Segmentation} & \multicolumn{2}{c}{Monocular Disparity Estimation} \\
			Method & IoU class & iIoU class & IoU cat & iIoU cat & AP & AP 50\% & AP 100m & AP 50m & Mean Error {[}$px${]} & RMS Error {[}$px${]} \\ \toprule % & Runtime [s]\\ \hline \hline
			\multicolumn{11}{c}{\textbf{Semantic segmentation, instance segmentation and depth regression methods (this work)}}\\ \midrule
			Multi-Task Learning & 78.5 & 57.4 & 89.9 & 77.7 & 21.6 & 39.0 & 35.0 & 37.0 & 2.92 & 5.88 \\ \midrule%& - \\
			\multicolumn{11}{c}{\textbf{Semantic segmentation and instance segmentation methods}}\\ \midrule
			Uhrig et al. \cite{uhrig2016pixel} & 64.3 & 41.6 & 85.9 & 73.9 & 8.9 & 21.1 & 15.3 & 16.7 &-&-\\ \midrule%& - \\ \hline
			\multicolumn{11}{c}{\textbf{Instance segmentation only methods}}\\ \midrule
			Mask R-CNN \cite{he2017mask} &-&-&-&-& 26.2 & 49.9 & 37.6 & 40.1 & -& -\\ %& 60.0 \\ \hline
            Deep Watershed \cite{bai2016deep} &-&-&-&-& 19.4 & 35.3 & 31.4 & 36.8 & -& -\\ %& 60.0 \\ \hline
            R-CNN + MCG \cite{Cordts2016Cityscapes} &-&-&-&-& 4.6 & 12.9 & 7.7 & 10.3 & -& -\\ \midrule %& 60.0 \\ \hline
			\multicolumn{11}{c}{\textbf{Semantic segmentation only methods}}\\ \midrule
			DeepLab V3 \cite{chen2017rethinking} & 81.3 & 60.9 & 91.6 & 81.7 &-&-&-&-&-&-\\%& 
			PSPNet \cite{zhao2016pyramid} & 81.2 & 59.6 & 91.2 & 79.2 &-&-&-&-&-&-\\%& 
			%LRR-4x \cite{ghiasi2016laplacian} & 71.8 & 47.9 & 88.4 & 73.9 &-&-&-&-&-&-\\%& 
			Adelaide \cite{lin2015efficient} & 71.6 & 51.7 & 87.3 & 74.1 &-&-&-&-&-&-\\%& 4.0\\
			%DeepLab \cite{chen2016deeplab} & 70.4 & 42.6 & 86.4 & 67.7 &-&-&-&-&-&-\\%& 4.0\\
			%Dilation \cite{YuKoltun2016} & 67.1 & 42.0 & 86.5 & 71.1 &-&-&-&-&-&-\\%& 4.0 \\
			%FCN 8s \cite{long2014fully} & 65.3 & 41.7 & 85.7 & 70.1 &-&-&-&-&-&-\\%& 0.5 \\
			%SegNet \cite{badrinarayanan2015segnet} & 57.0 & 32.0 & 79.1 & 61.9 &-&-&-&-&-&-\\%& 0.06 \\
            \bottomrule
		\end{tabular}}
	\caption{\textbf{CityScapes Benchmark \cite{Cordts2016Cityscapes}.} We show results from the test dataset using the full resolution of $1024\times2048$ pixels. For the full leaderboard, please see \url{www.cityscapes-dataset.com/benchmarks}. The disparity (inverse depth) metrics were computed against the CityScapes depth maps, which are sparse and computed using SGM stereo \cite{hirschmuller2005accurate}. Note, these comparisons are not entirely fair, as many methods use ensembles of different training datasets. Our method is the first to address all three tasks with a single model.}
	\label{tbl:benchmarks}
\end{table*}

\section{Conclusions}

We have shown that correctly weighting loss terms is of paramount importance for multi-task learning problems. We demonstrated that homoscedastic (task) uncertainty is an effective way to weight losses. We derived a principled loss function which can learn a relative weighting automatically from the data and is robust to the weight initialization. We showed that this can improve performance for scene understanding tasks with a unified architecture for semantic segmentation, instance segmentation and per-pixel depth regression. We demonstrated modelling task-dependent homoscedastic uncertainty improves the model's representation and each task's performance when compared to separate models trained on each task individually.

There are many interesting questions left unanswered. Firstly, our results show that there is usually not a single optimal weighting for all tasks. Therefore, what is the optimal weighting? Is multitask learning is an ill-posed optimisation problem without a single higher-level goal?

A second interesting question is where the optimal location is for splitting the shared encoder network into separate decoders for each task? And, what network depth is best for the shared multi-task representation?

Finally, why do the semantics and depth tasks out-perform the semantics and instance tasks results in \tbl{multitasks}? Clearly the three tasks explored in this paper are complimentary and useful for learning a rich representation about the scene. It would be beneficial to be able to quantify the relationship between tasks and how useful they would be for multitask representation learning.

{\small
\bibliographystyle{ieee}
\bibliography{bib}
}

\clearpage
% \onecolumn
\appendix

\section{Model Architecture Details}
\label{apdx:arch}

We base our model on the recently introduced DeepLabV3 \cite{chen2017rethinking} segmentation architecture.
%We use \cite{wideresnet} as our base feature encoder (pretrained by \cite{bulo2017place}), with dilated convolutions, resulting in a feature map which is downsampled by a factor of 8 compared with the original input image. 
We use ResNet101 \cite{he2015deep} as our base feature encoder, with dilated convolutions, resulting in a feature map which is downsampled by a factor of 8 compared with the original input image. 
We then append dilated (atrous) convolutional ASPP module \cite{chen2017rethinking}. This module is designed to improve the contextual reasoning of the network. We use an ASPP module comprised of four parallel convolutional layers, with 256 output channels and dilation rates (1, 12, 24, 36), with kernel sizes ($1^2$, $3^2$, $3^2$, $3^2$). Additionally, we also apply global average pooling to the encoded features, and convolve them to 256 dimensions with a $1 \times 1$ kernel. We apply batch normalisation to each of these layers and concatenate the resulting 1280 features together. This produces the shared representation between each task.

We then split the network, to decode this representation to a given task output. For each task, we construct a decoder consisting of two layers. First, we apply a $1 \times 1$ convolution, outputting 256 features, followed by batch normalisation and a non-linear activation. Finally, we convolve this output to the required dimensions for a given task. For classification, this will be equal to the number of semantic classes, otherwise the output will be 1 or 2 channels for depth or instance segmentation respectively. Finally, we apply bilinear upsampling
to scale the output to the same resolution as the input.

The majority of the model's parameters and depth is in the feature encoding, with very little flexibility in each task decoder. This illustrates the attraction of multitask learning; most of the compute can be shared between each task to learn a better shared representation.

%We apply in-place activated batch normalisation \cite{bulo2017place} to our model. In order to train with a sufficient crop size and batch size, in-place activated batch normalisation enables us to share the memory between the normalisation and non-linearity layers, throughout the network. It does this by considering the activation and batch normalisation layers as invertible. This requires for us to use an invertible non-linearity, as standard ReLU is not invertible. We apply in-place activated batch normalisation to all layers of our model, using a leaky ReLU activation, with slope $0.01$.

\subsection{Optimisation}

For all experiments, we use an initial learning rate of $2.5\times10^{-3}$ and polynomial learning rate decay $(1-\frac{iter}{max~iter})^{0.9}$. We train using stochastic gradient descent, with Nesterov updates and momentum $0.9$ and weight decay $10^{−4}$. We conduct all experiments in this paper using PyTorch.

For the experiments on the Tiny CityScapes validation dataset (using a down-sampled resolution of $128\times256$) we train over $50,000$ iterations, using $256 \times 256$ crops with batch size of 8 on a single NVIDIA 1080Ti GPU. We apply random horizontal flipping to the data.

For the full-scale CityScapes benchmark experiment, we train over $100,000$ iterations with a batch size of 16. We apply random horizontal flipping (with probability 0.5) and random scaling (selected from 0.7 - 2.0) to the data during training, before making a $512 \times 512$ crop. The training data is sampled uniformly, and is randomly shuffled for each epoch. Training takes five days on a single computer with four NVIDIA 1080Ti GPUs.

\section{Further Analysis}
\label{apdx:analysis}

\begin{figure*}[t]
\begin{subfigure}[t]{0.32\linewidth}
\begin{center}
  \includegraphics[width=\linewidth]{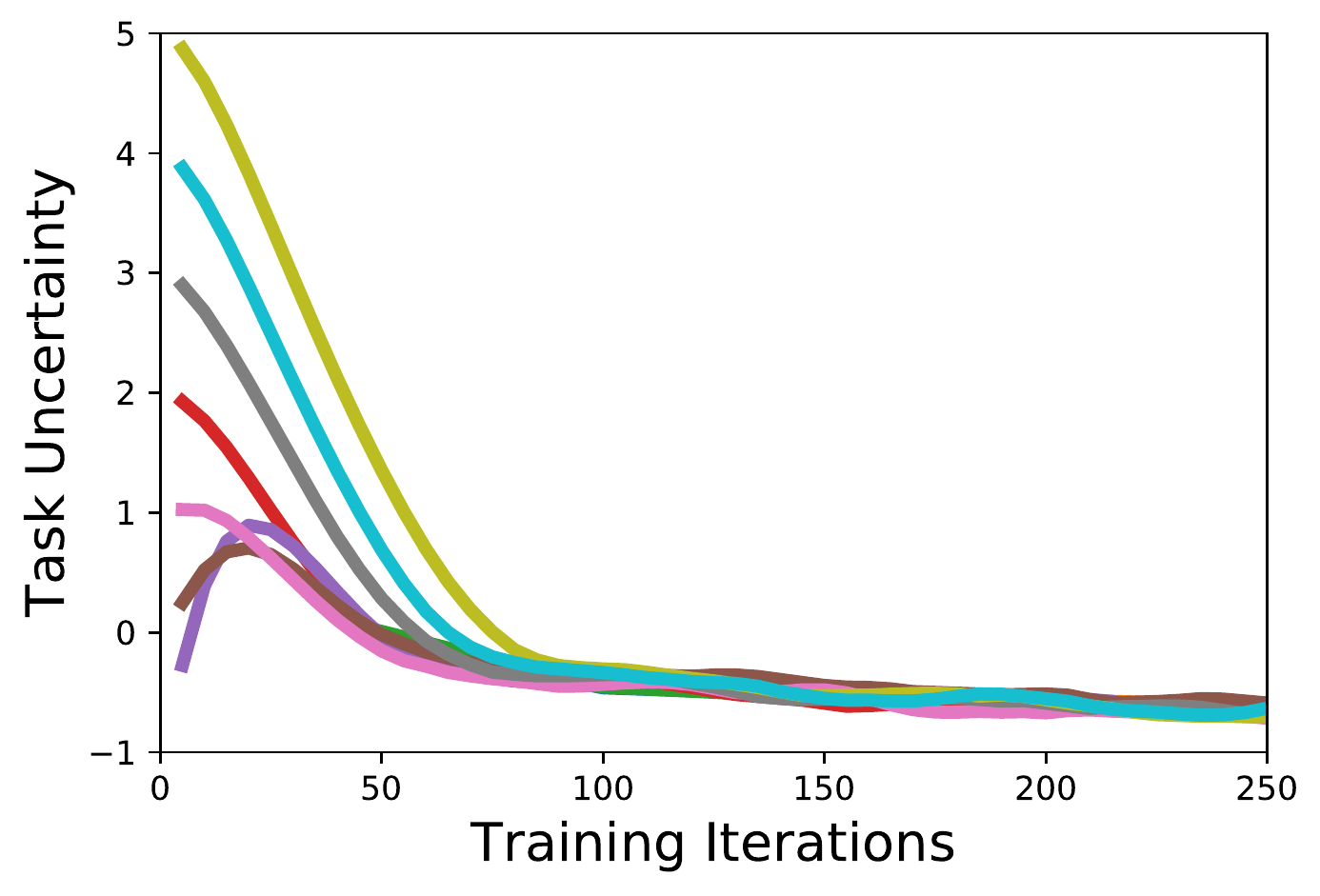}
  \caption{Semantic segmentation task}
\end{center}
\end{subfigure}
\begin{subfigure}[t]{0.32\linewidth}
\begin{center}
  \includegraphics[width=\linewidth]{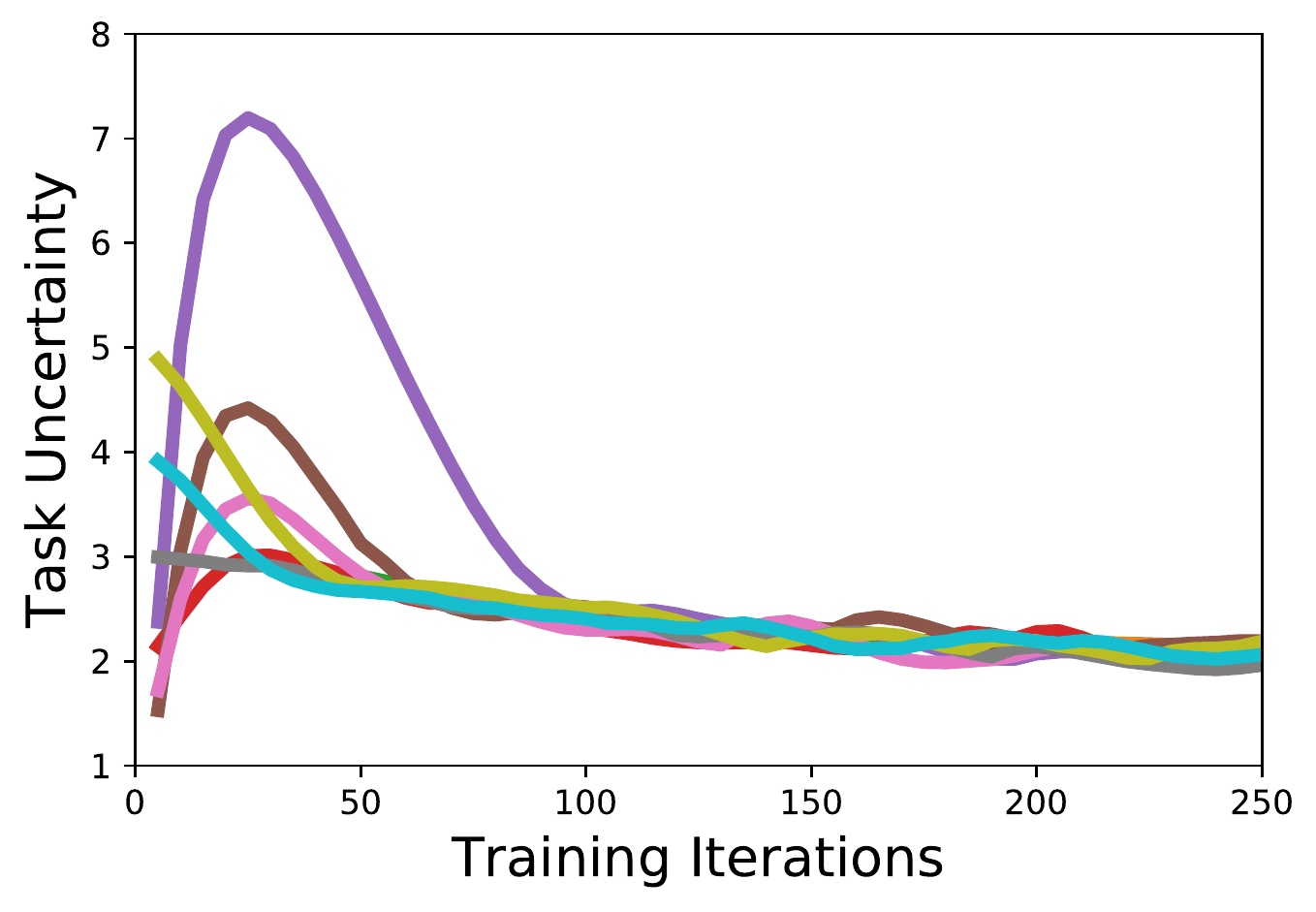}
  \caption{Instance segmentation task}
\end{center}
\end{subfigure}
\begin{subfigure}[t]{0.32\linewidth}
\begin{center}
  \includegraphics[width=\linewidth]{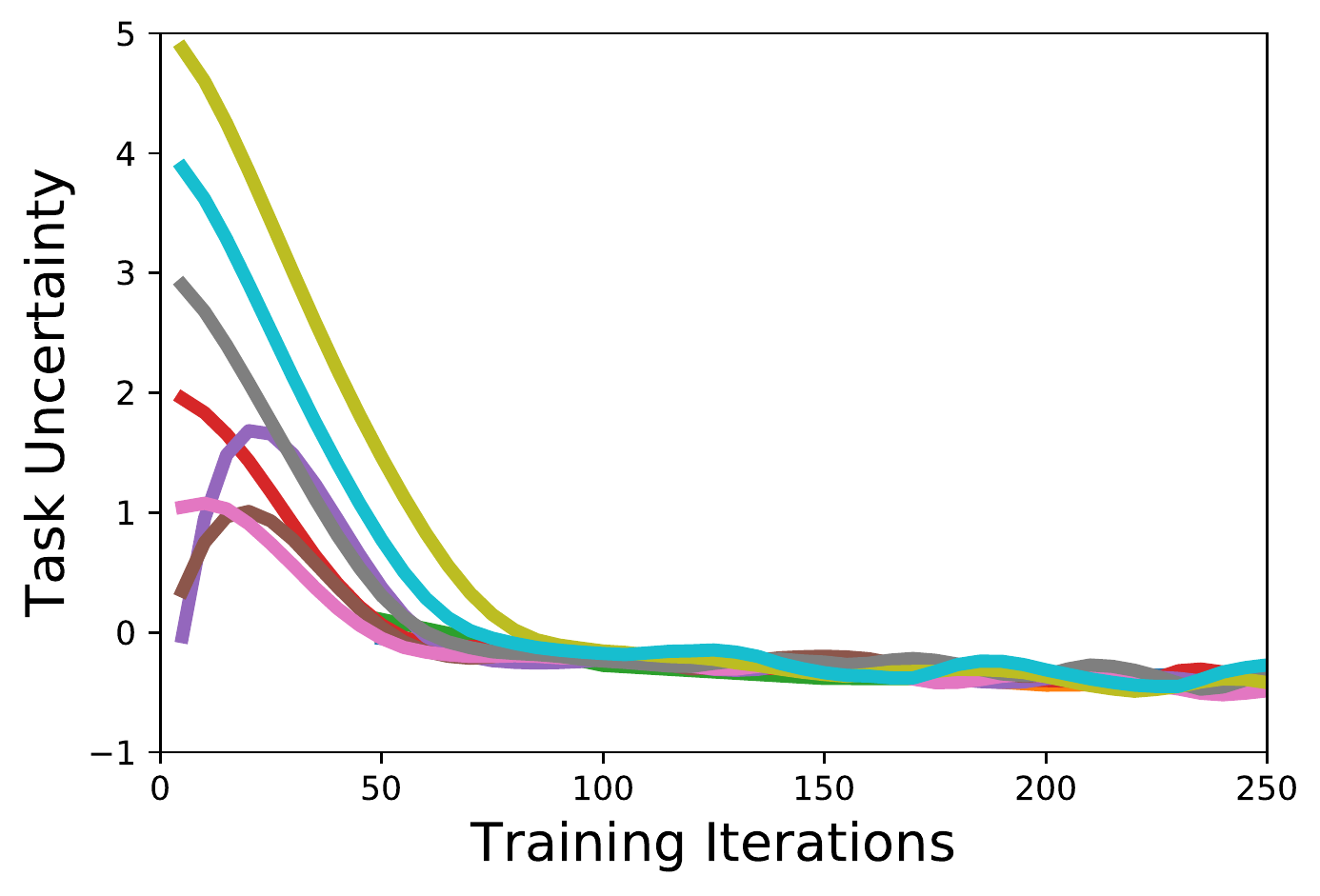}
  \caption{Depth regression task}
\end{center}
\end{subfigure}
   \caption{\textbf{Training plots showing convergence of homoscedastic noise and task loss} for an array of initialisation choices for the homoscedastic uncertainty terms for all three tasks. Each plot shows the the homoscedastic noise value optimises to the same solution from a variety of initialisations. Despite the network taking $10,000+$ iterations for the training loss to converge, the task uncertainty converges very rapidly after only $~100$ iterations.}
\label{fig:convergence}
\end{figure*}

This task uncertainty loss is also robust to the value we use to initialise the task uncertainty values.
One of the attractive properties of our approach to weighting multi-task losses is that it is robust to the initialisation choice for the homoscedastic noise parameters. \fig{convergence} shows that for an array of initial choices of $\log \sigma^2$ from $-2.0$ to $5.0$ the homoscedastic noise and task loss is able to converge to the same minima. Additionally, the homoscedastic noise terms converges after only $~100$ iterations, while the network requires $30,000+$ iterations to train. Therefore our model is robust to the choice of initial value for the weighting terms.

\begin{figure*}[t]
\begin{center}
\begin{subfigure}[t]{0.32\linewidth}
\begin{center}
  \includegraphics[width=\linewidth]{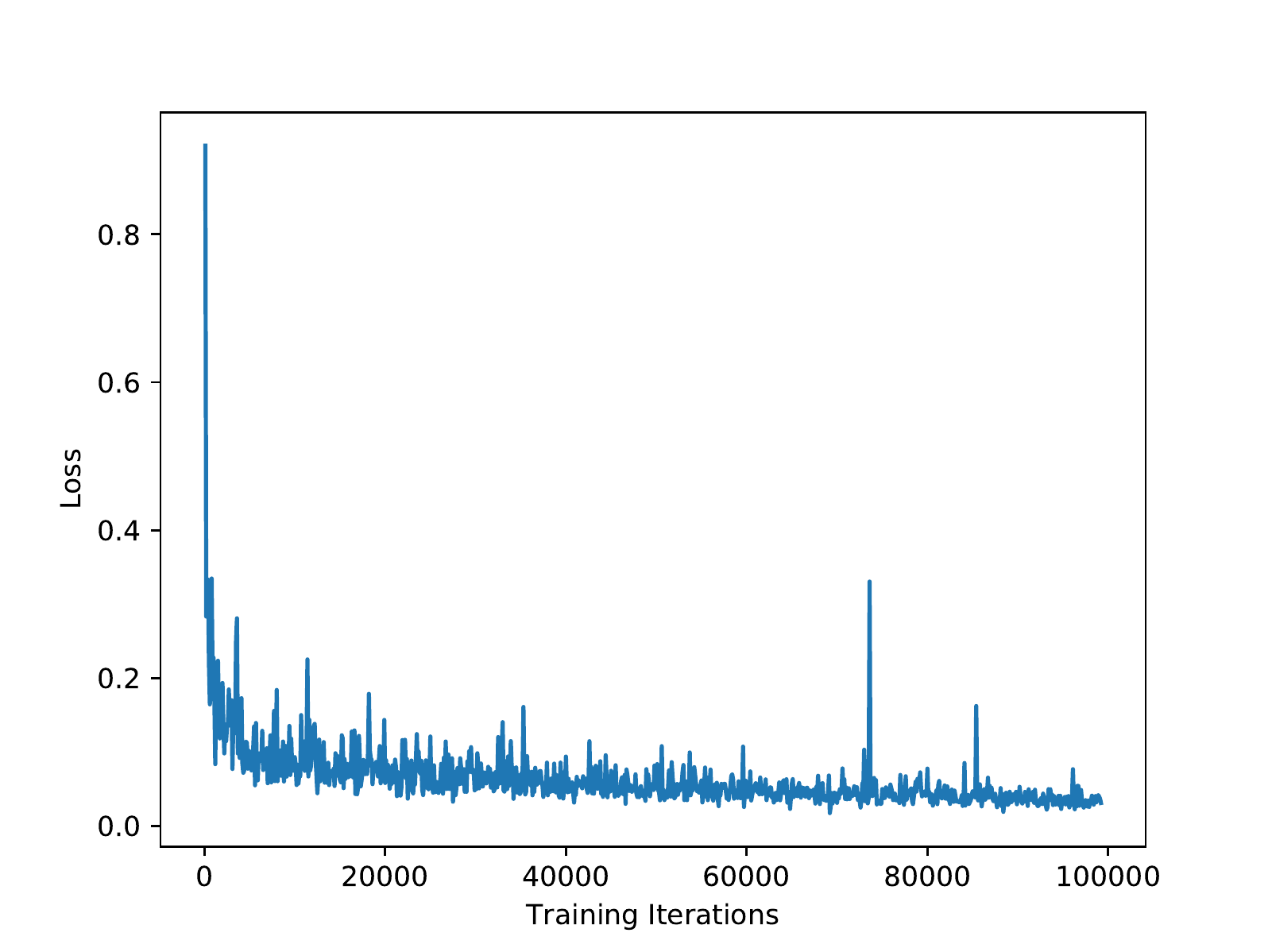}
\end{center}
\end{subfigure}
\begin{subfigure}[t]{0.32\linewidth}
\begin{center}
  \includegraphics[width=\linewidth]{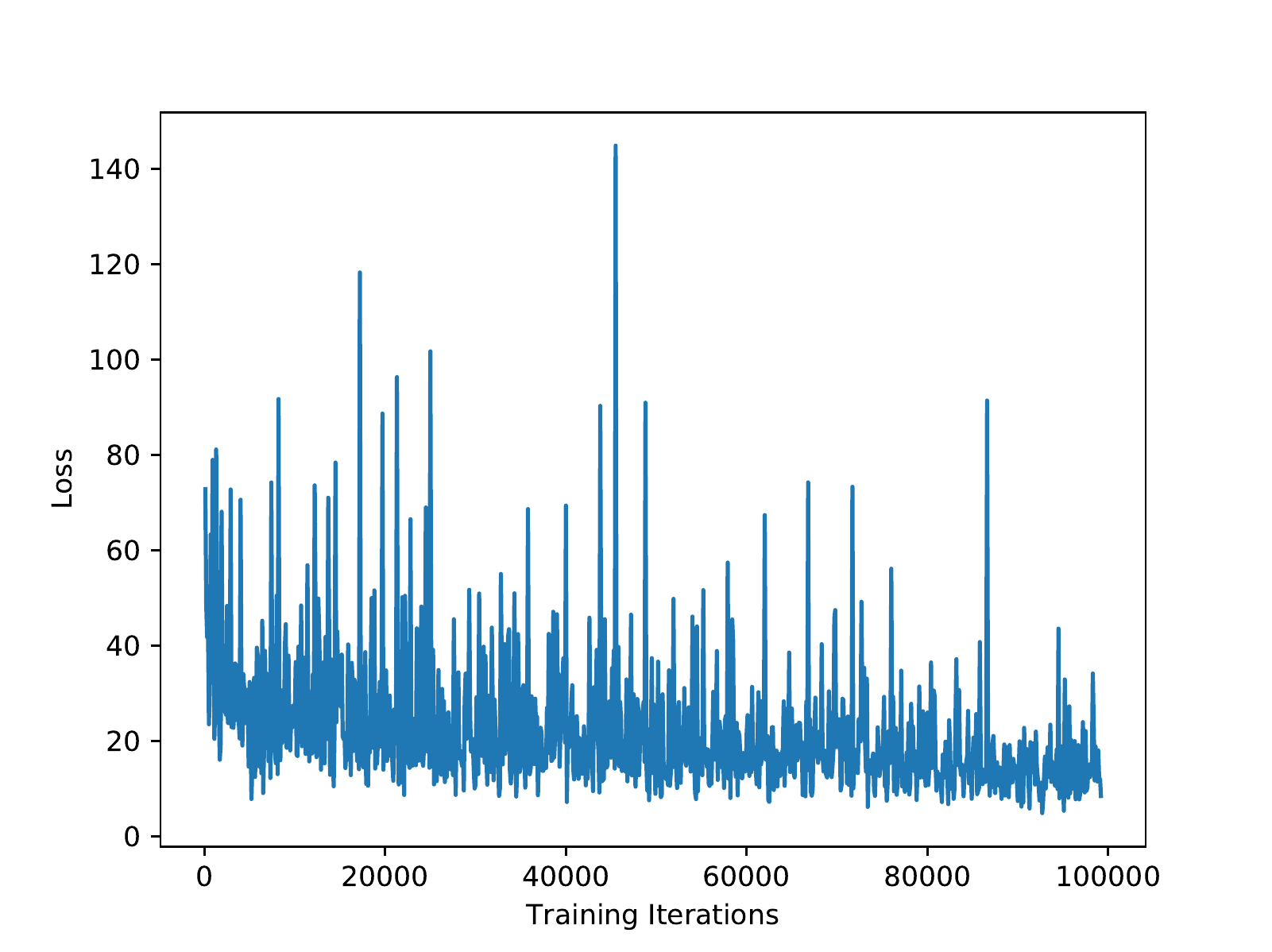}
\end{center}
\end{subfigure}
\begin{subfigure}[t]{0.32\linewidth}
\begin{center}
  \includegraphics[width=\linewidth]{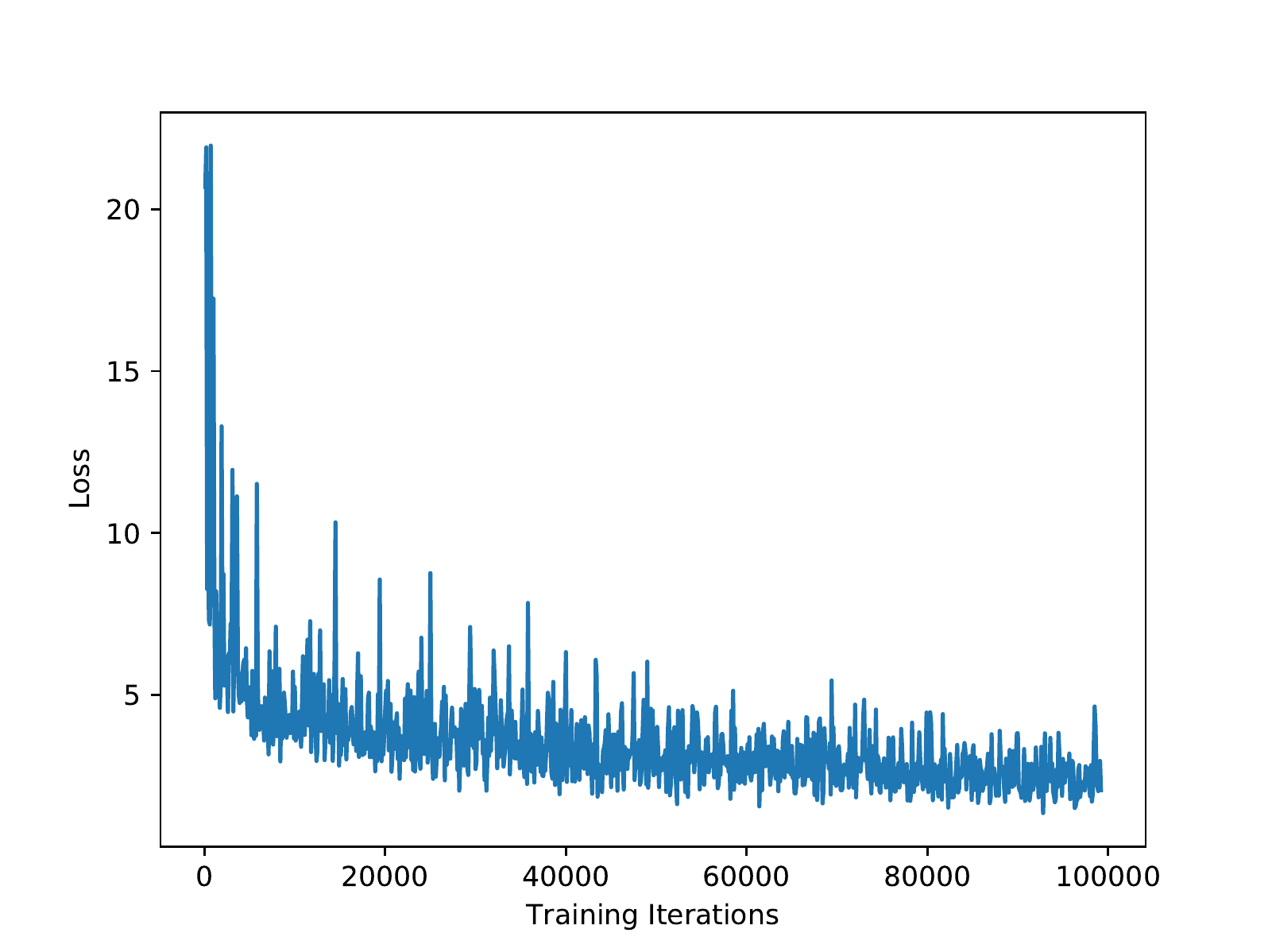}
\end{center}
\end{subfigure}

\begin{subfigure}[t]{0.32\linewidth}
\begin{center}
  \includegraphics[width=\linewidth]{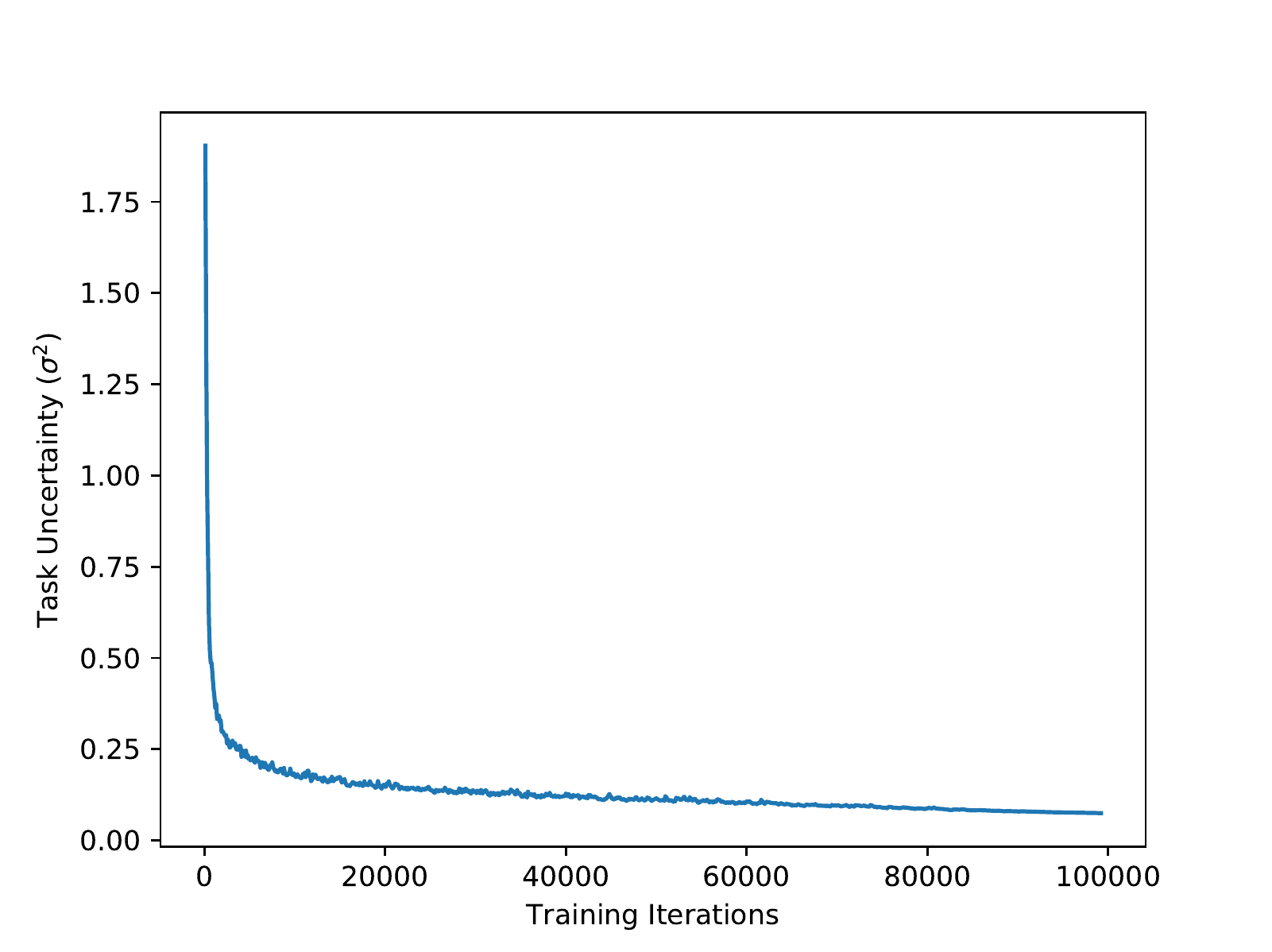}
  \caption{Semantic segmentation task}
\end{center}
\end{subfigure}
\begin{subfigure}[t]{0.32\linewidth}
\begin{center}
  \includegraphics[width=\linewidth]{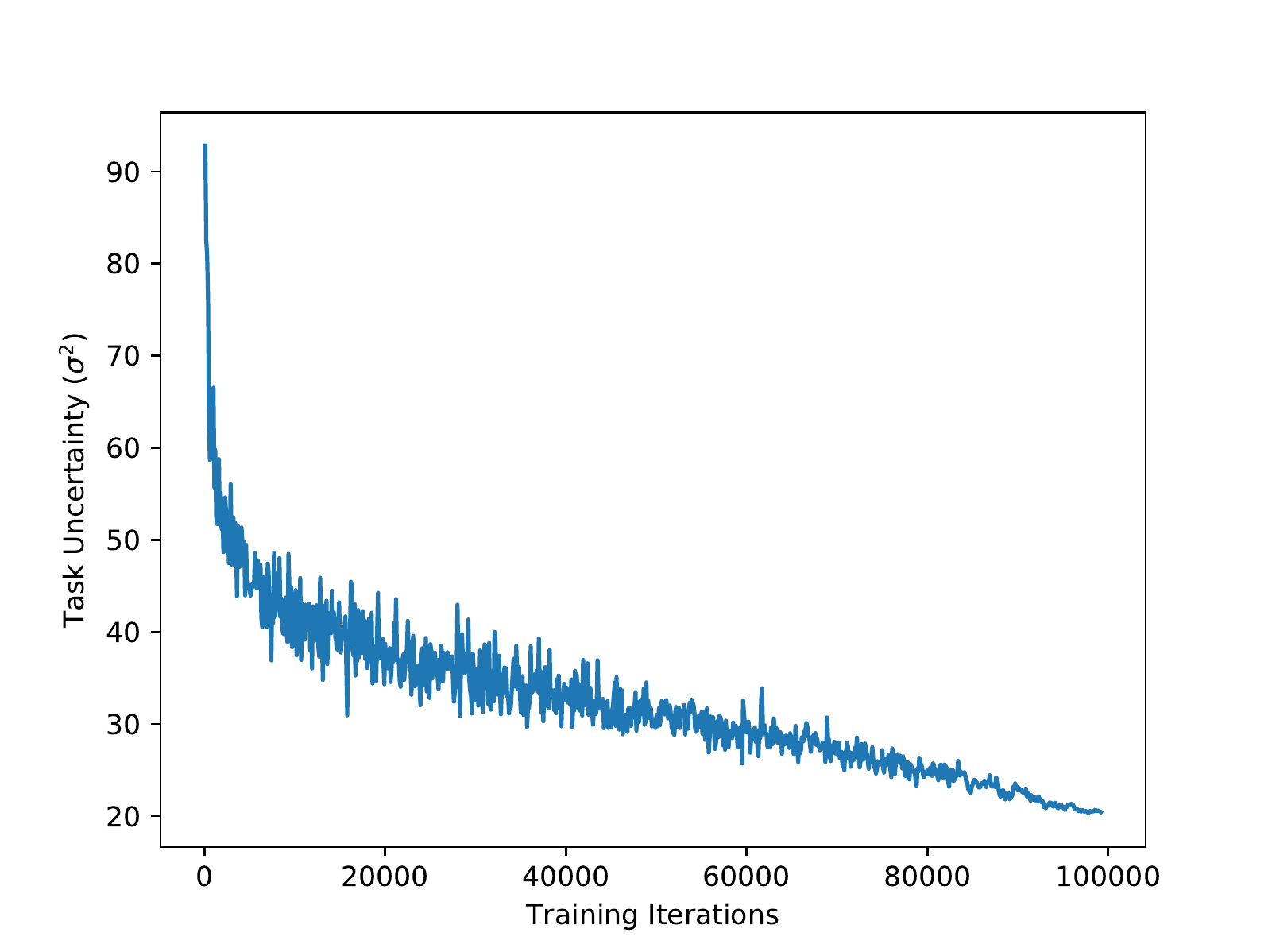}
  \caption{Instance segmentation task}
\end{center}
\end{subfigure}
\begin{subfigure}[t]{0.32\linewidth}
\begin{center}
  \includegraphics[width=\linewidth]{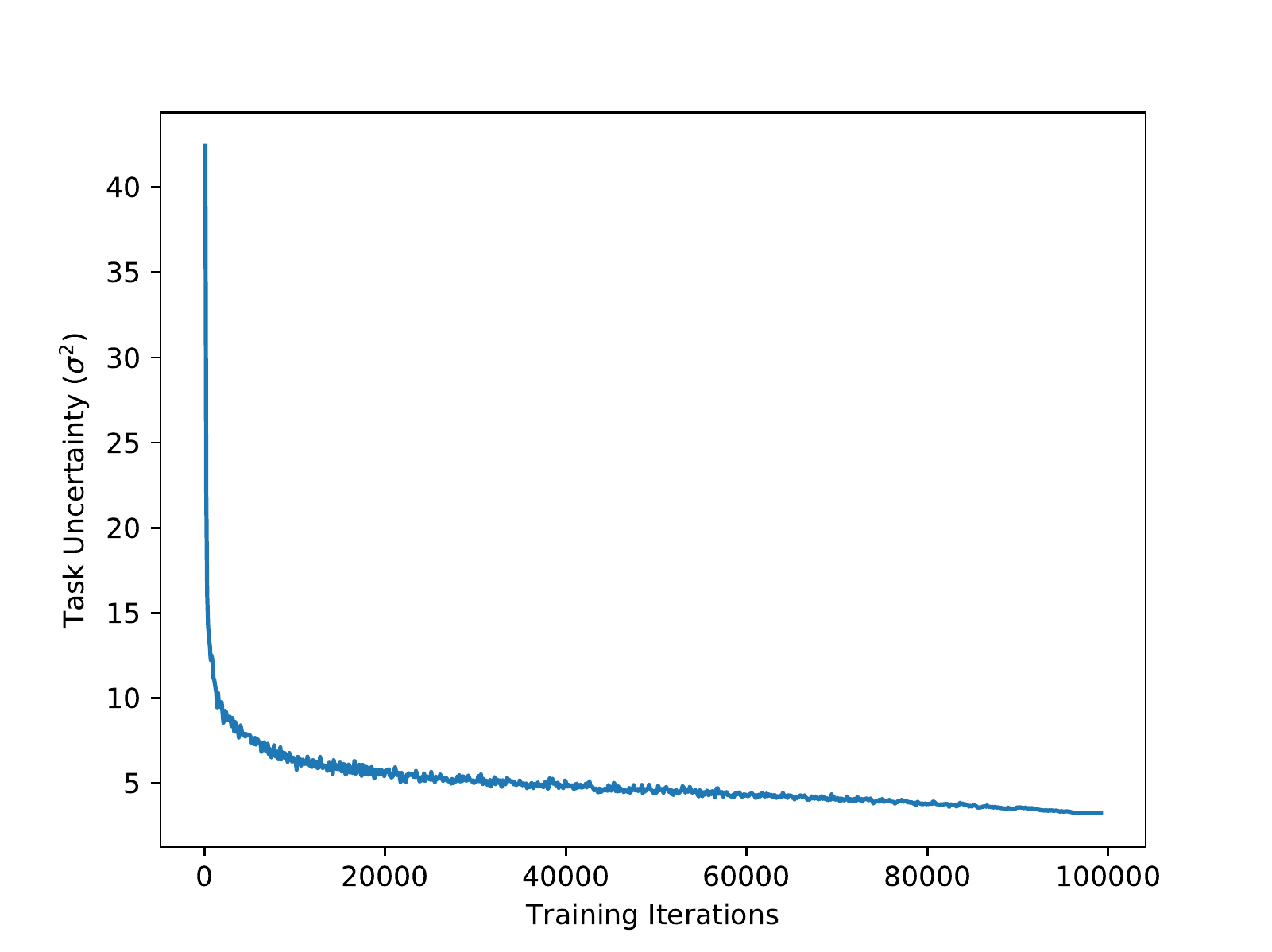}
  \caption{Depth regression task}
\end{center}
\end{subfigure}
\end{center}
   \caption{\textbf{Learning task uncertainty.} These training plots show the losses and task uncertainty estimates for each task during training. Results are shown for the final model, trained on the fullsize CityScapes dataset.}
\label{fig:taskunc}
\end{figure*}

\fig{taskunc} shows losses and uncertainty estimates for each task during training of the final model on the full-size CityScapes dataset. At a point 500 iterations into training, the model estimates task variance of 0.60, 62.5 and 13.5 for semantic segmentation, instance segmentation and depth regression, respectively. Becuase the losses are weighted by the inverse of the uncertainty estimates, this results in a task weighting ratio of approximately 23 : 0.22 : 1 between semantics, instance and depth, respectively. At the conclusion of training, the three tasks have uncertainty estimates of 0.075, 3.25 and 20.4, which results in effective weighting between the tasks of 43: 0.16 : 1. This shows how the task uncertainty estimates evolve over time, and the approximate final weightings the network learns. We observe they are far from uniform, as is often assumed in previous literature.

Interestingly, we observe that this loss allows the network to dynamically tune the weighting. Typically, the homoscedastic noise terms decrease in magnitude as training progresses. This makes sense, as during training the model becomes more effective at a task. Therefore the error, and uncertainty, will decrease. This has a side-effect of increasing the effective learning rate -- because the overall uncertainty decreases, the weight for each task's loss increases. In our experiments we compensate for this by annealing the learning rate with a power law.

\begin{figure*}[t]
\section{Further Qualitative Results}
\label{apdx:qual}
\makebox[\textwidth][c]{
\resizebox{\linewidth}{!}{
\begin{subfigure}[t]{0.24\linewidth}
\begin{center}
		\includegraphics[width=\linewidth,trim={0px 60px 0 0px},clip]{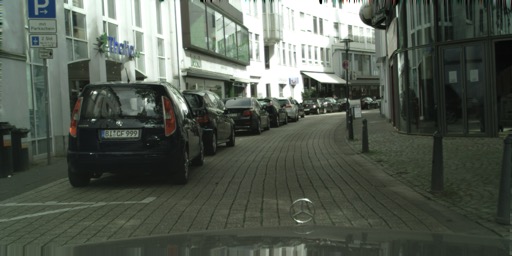}
		\includegraphics[width=\linewidth,trim={0px 60px 0 0px},clip]{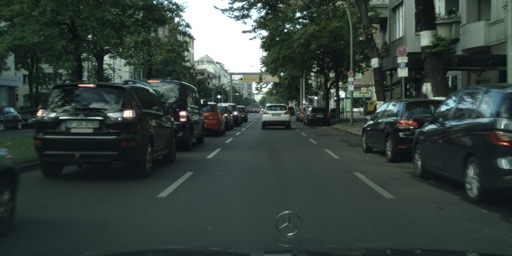}
		\includegraphics[width=\linewidth,trim={0px 60px 0 0px},clip]{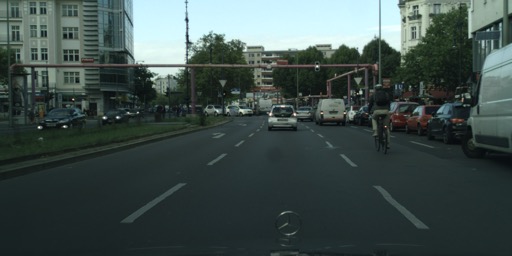}
		\includegraphics[width=\linewidth,trim={0px 60px 0 0px},clip]{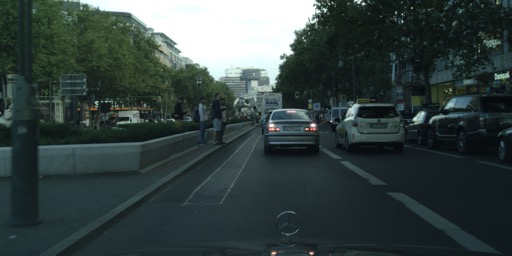}
		\includegraphics[width=\linewidth,trim={0px 60px 0 0px},clip]{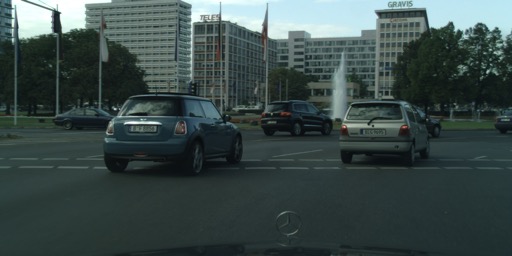}
		\includegraphics[width=\linewidth,trim={0px 60px 0 0px},clip]{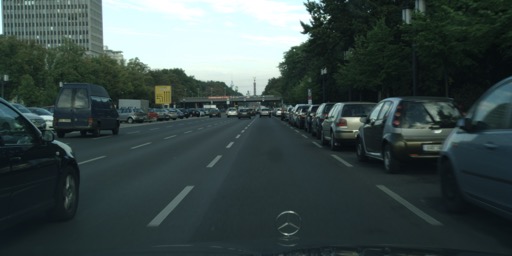}
		\includegraphics[width=\linewidth,trim={0px 60px 0 0px},clip]{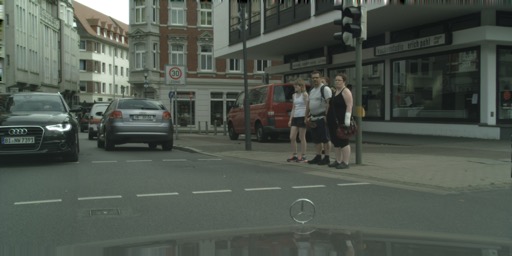}
		\includegraphics[width=\linewidth,trim={0px 60px 0 0px},clip]{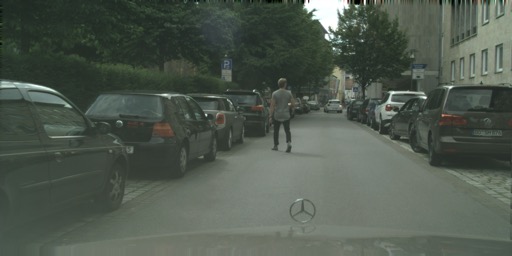}
		\includegraphics[width=\linewidth,trim={0px 60px 0 0px},clip]{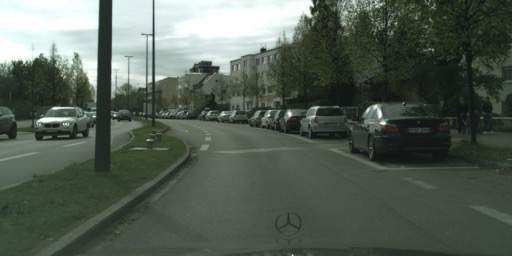}
		\includegraphics[width=\linewidth,trim={0px 60px 0 0px},clip]{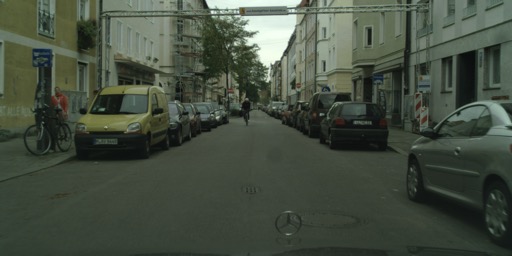}
		\includegraphics[width=\linewidth,trim={0px 60px 0 0px},clip]{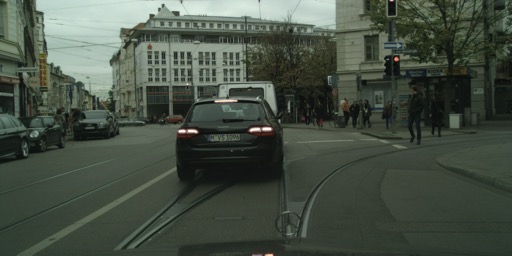}
  \caption{Input image}
\end{center}
\end{subfigure}
\begin{subfigure}[t]{0.24\linewidth}
\begin{center}
		\includegraphics[width=\linewidth,trim={0px 60px 0 0px},clip]{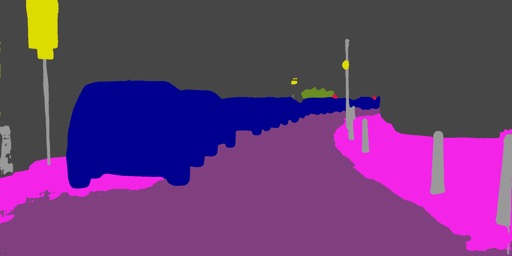}
		\includegraphics[width=\linewidth,trim={0px 60px 0 0px},clip]{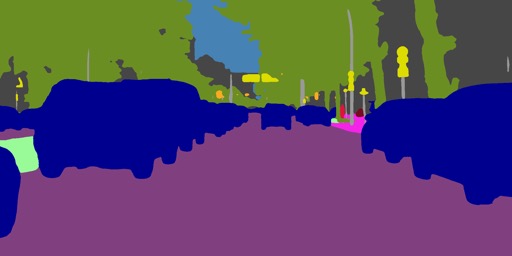}
		\includegraphics[width=\linewidth,trim={0px 60px 0 0px},clip]{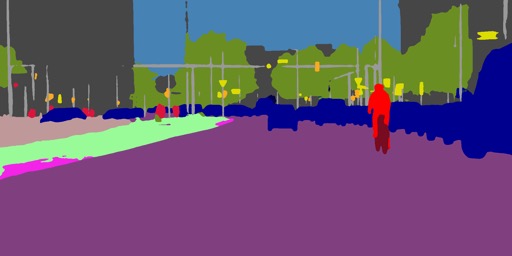}
		\includegraphics[width=\linewidth,trim={0px 60px 0 0px},clip]{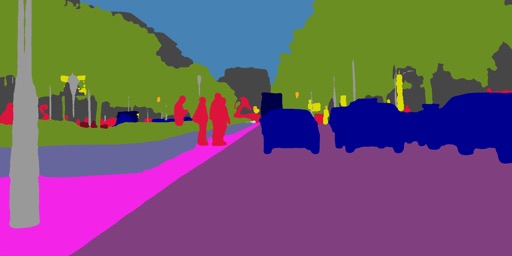}
		\includegraphics[width=\linewidth,trim={0px 60px 0 0px},clip]{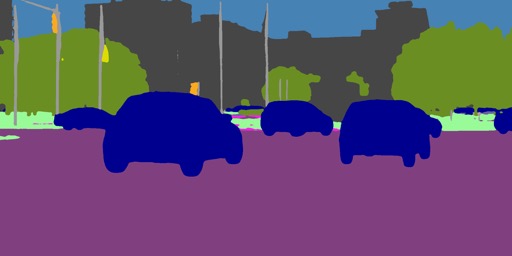}
		\includegraphics[width=\linewidth,trim={0px 60px 0 0px},clip]{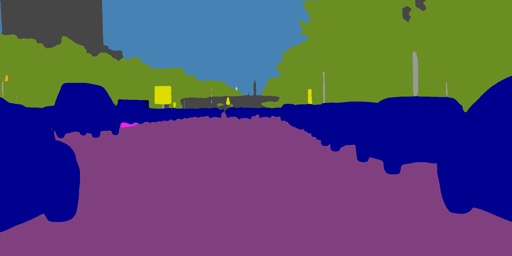}
		\includegraphics[width=\linewidth,trim={0px 60px 0 0px},clip]{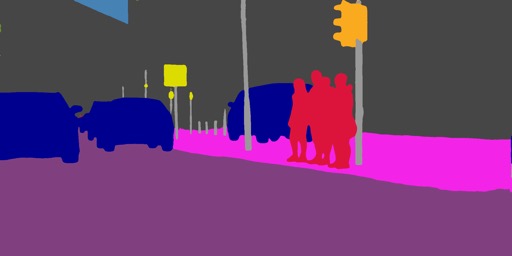}
		\includegraphics[width=\linewidth,trim={0px 60px 0 0px},clip]{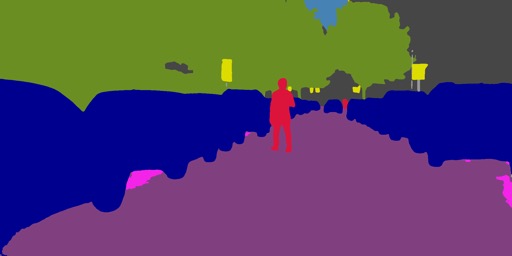}
		\includegraphics[width=\linewidth,trim={0px 60px 0 0px},clip]{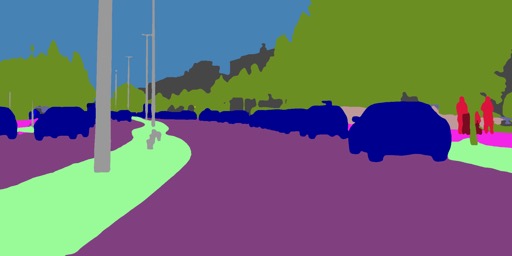}
		\includegraphics[width=\linewidth,trim={0px 60px 0 0px},clip]{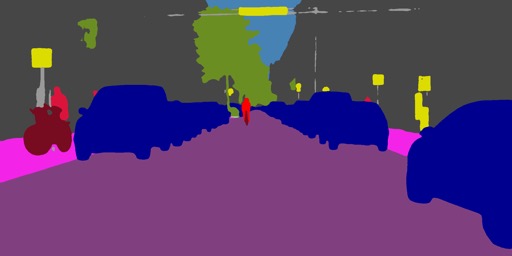}
		\includegraphics[width=\linewidth,trim={0px 60px 0 0px},clip]{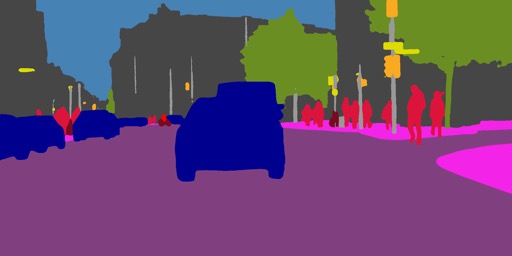}
  \caption{Semantic segmentation}
\end{center}
\end{subfigure}
\begin{subfigure}[t]{0.24\linewidth}
\begin{center}
		\includegraphics[width=\linewidth,trim={0px 60px 0 0px},clip]{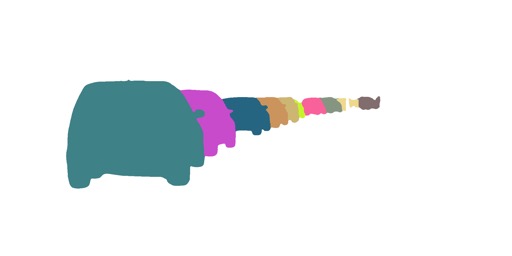}
		\includegraphics[width=\linewidth,trim={0px 60px 0 0px},clip]{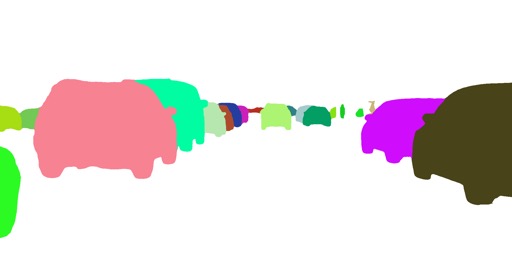}
		\includegraphics[width=\linewidth,trim={0px 60px 0 0px},clip]{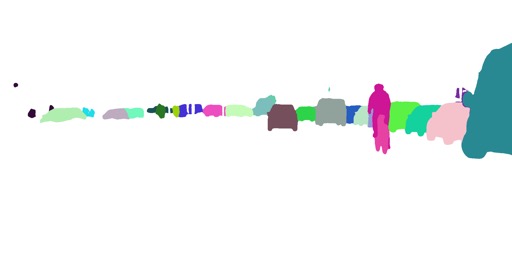}
		\includegraphics[width=\linewidth,trim={0px 60px 0 0px},clip]{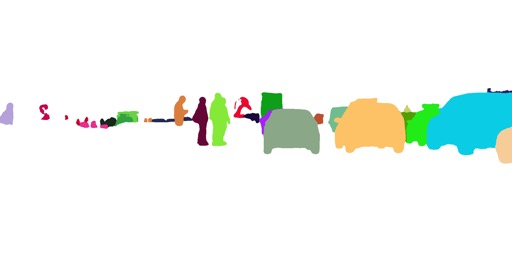}
		\includegraphics[width=\linewidth,trim={0px 60px 0 0px},clip]{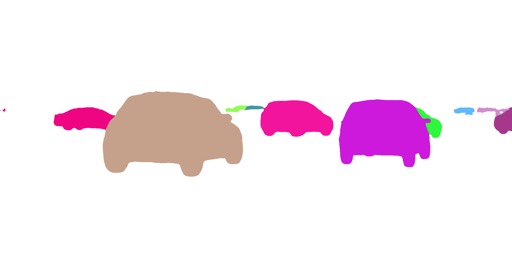}
		\includegraphics[width=\linewidth,trim={0px 60px 0 0px},clip]{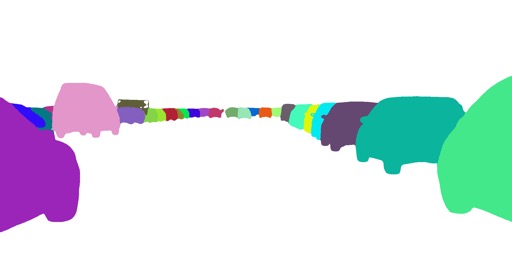}
		\includegraphics[width=\linewidth,trim={0px 60px 0 0px},clip]{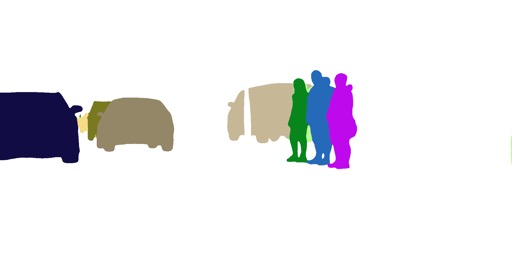}
		\includegraphics[width=\linewidth,trim={0px 60px 0 0px},clip]{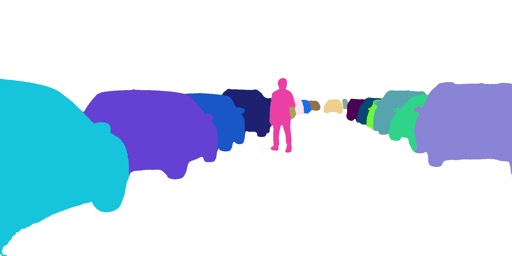}
		\includegraphics[width=\linewidth,trim={0px 60px 0 0px},clip]{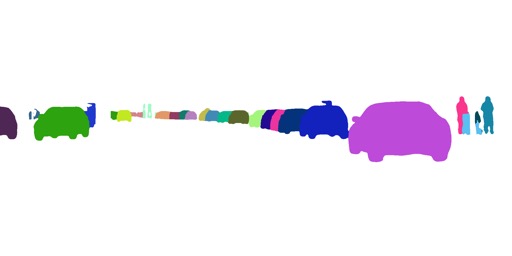}
		\includegraphics[width=\linewidth,trim={0px 60px 0 0px},clip]{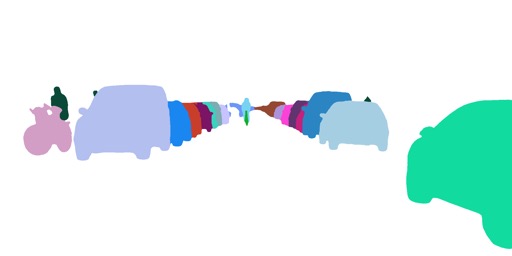}
		\includegraphics[width=\linewidth,trim={0px 60px 0 0px},clip]{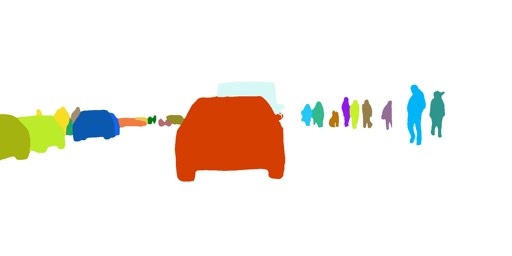}
  \caption{Instance segmentation}
\end{center}
\end{subfigure}
\begin{subfigure}[t]{0.24\linewidth}
\begin{center}
		\includegraphics[width=\linewidth,trim={0px 60px 0 0px},clip]{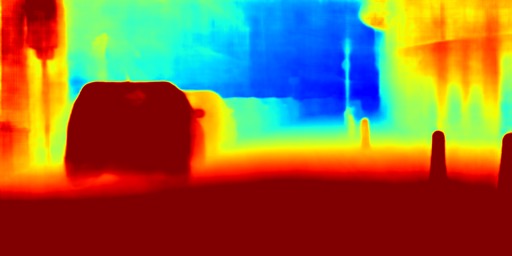}
		\includegraphics[width=\linewidth,trim={0px 60px 0 0px},clip]{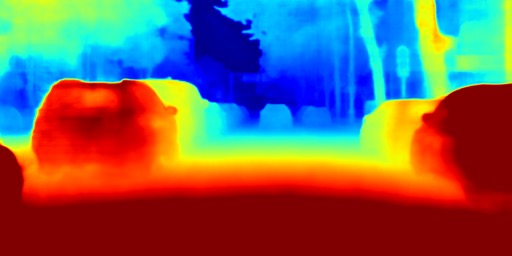}
		\includegraphics[width=\linewidth,trim={0px 60px 0 0px},clip]{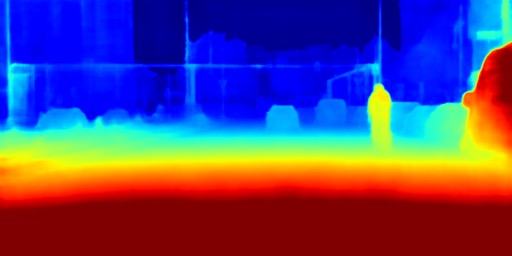}
		\includegraphics[width=\linewidth,trim={0px 60px 0 0px},clip]{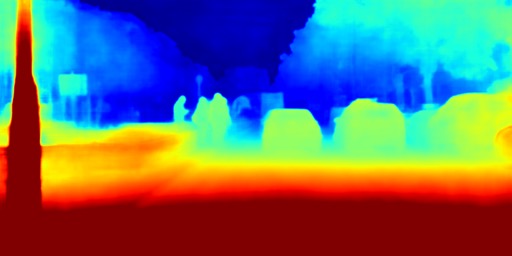}
		\includegraphics[width=\linewidth,trim={0px 60px 0 0px},clip]{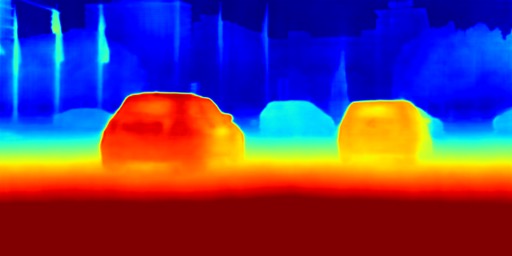}
		\includegraphics[width=\linewidth,trim={0px 60px 0 0px},clip]{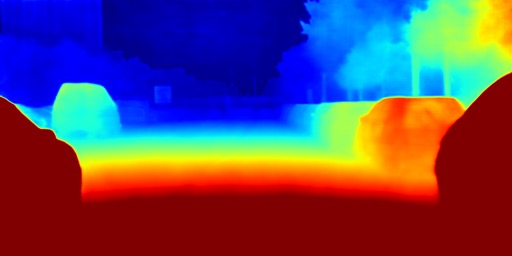}
		\includegraphics[width=\linewidth,trim={0px 60px 0 0px},clip]{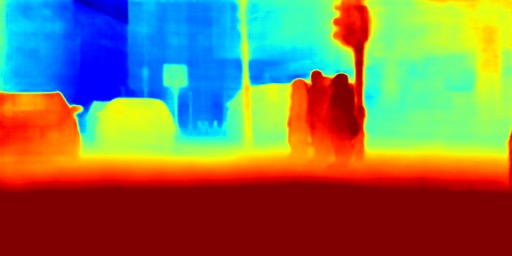}
		\includegraphics[width=\linewidth,trim={0px 60px 0 0px},clip]{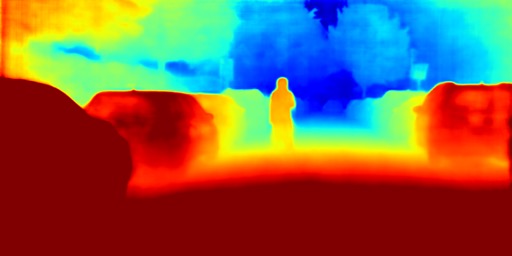}
		\includegraphics[width=\linewidth,trim={0px 60px 0 0px},clip]{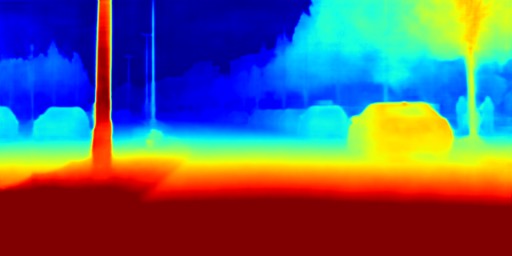}
		\includegraphics[width=\linewidth,trim={0px 60px 0 0px},clip]{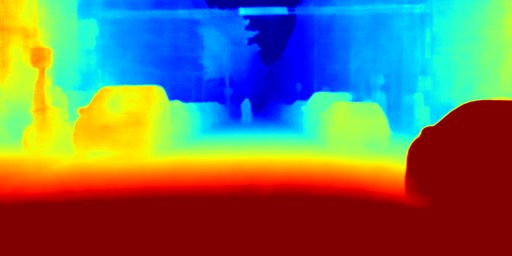}
		\includegraphics[width=\linewidth,trim={0px 60px 0 0px},clip]{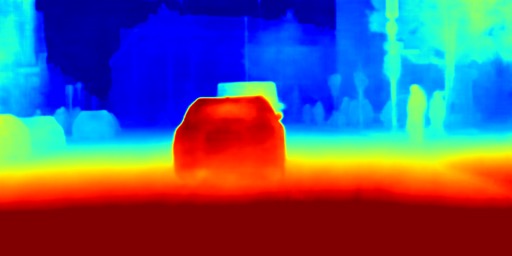}
  \caption{Depth regression}
\end{center}
\end{subfigure}}}
	\caption{\textbf{More qualitative results on test images from the CityScapes dataset.}}
	\label{fig:cityscapesquallarge}
\end{figure*}

Finally, a comment on the model's failure modes. The model exhibits similar failure modes to state-of-the-art single-task models. For example, failure with objects out of the training distribution, occlusion or visually challenging situations. However, we also observe our multi-task model tends to fail with similar effect in all three modalities. Ie. an erroneous pixel's prediction in one task will often be highly correlated with error in another modality. Some examples can be seen in \fig{cityscapesquallarge}.

\begin{figure*}[t]
\section{Failure Examples}
\label{apdx:qual}
\makebox[\textwidth][c]{
\resizebox{\linewidth}{!}{
\begin{subfigure}[t]{0.24\linewidth}
\begin{center}
		\includegraphics[width=\linewidth,trim={0px 60px 0 0px},clip]{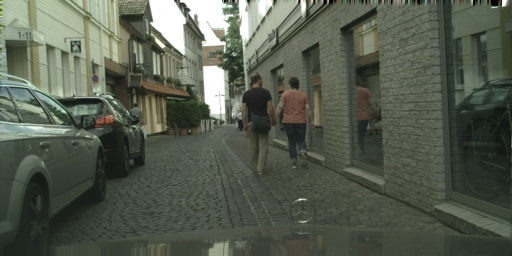}
		\includegraphics[width=\linewidth,trim={0px 60px 0 0px},clip]{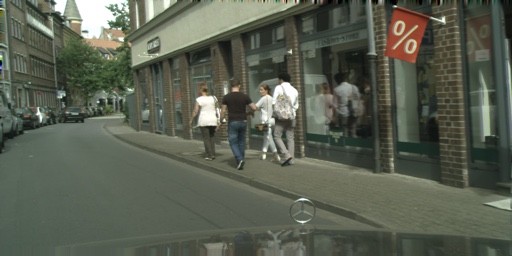}
		\includegraphics[width=\linewidth,trim={0px 60px 0 0px},clip]{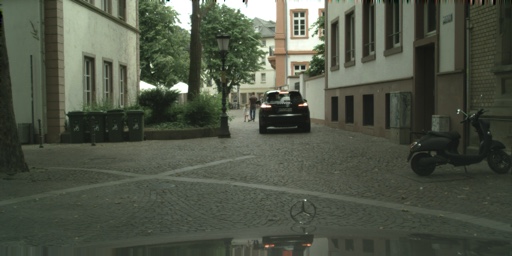}
		\includegraphics[width=\linewidth,trim={0px 60px 0 0px},clip]{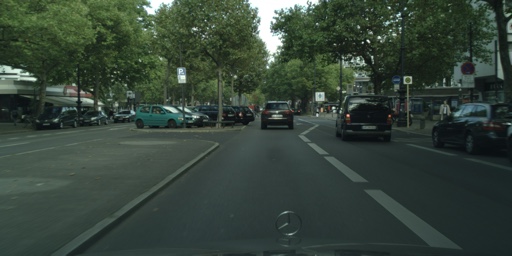}
		\includegraphics[width=\linewidth,trim={0px 60px 0 0px},clip]{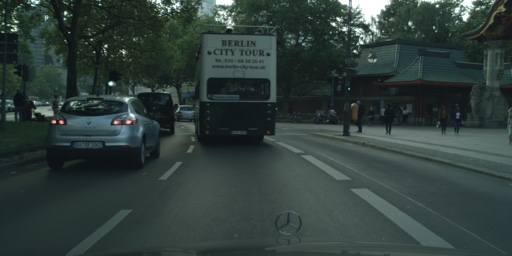}
		\includegraphics[width=\linewidth,trim={0px 60px 0 0px},clip]{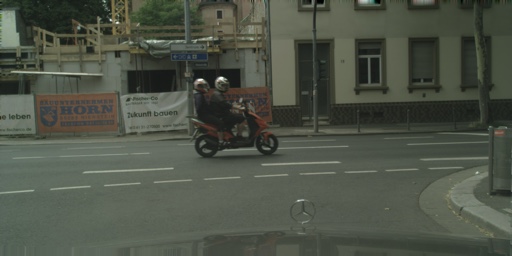}
		\includegraphics[width=\linewidth,trim={0px 60px 0 0px},clip]{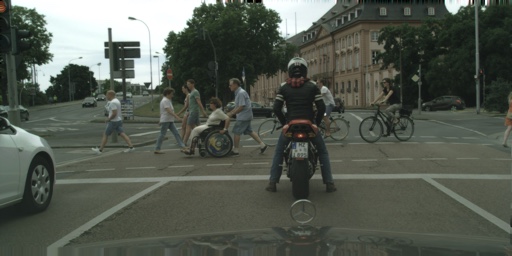}
		\includegraphics[width=\linewidth,trim={0px 60px 0 0px},clip]{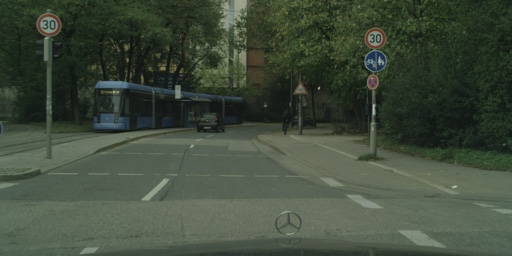}
		\includegraphics[width=\linewidth,trim={0px 60px 0 0px},clip]{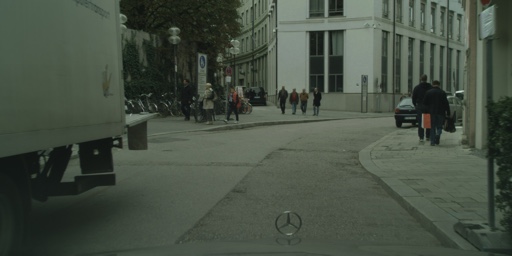}
		\includegraphics[width=\linewidth,trim={0px 60px 0 0px},clip]{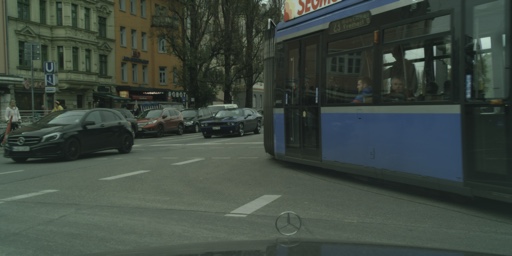}
  \caption{Input image}
\end{center}
\end{subfigure}
\begin{subfigure}[t]{0.24\linewidth}
\begin{center}
		\includegraphics[width=\linewidth,trim={0px 60px 0 0px},clip]{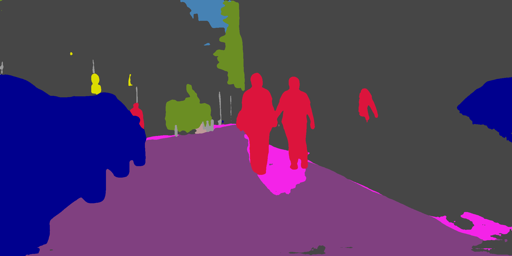}
		\includegraphics[width=\linewidth,trim={0px 60px 0 0px},clip]{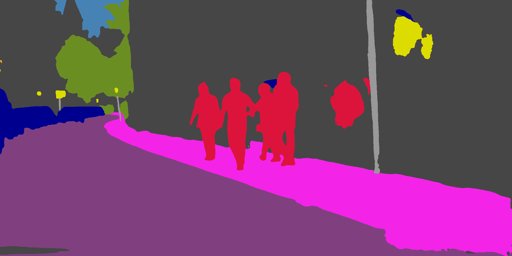}
		\includegraphics[width=\linewidth,trim={0px 60px 0 0px},clip]{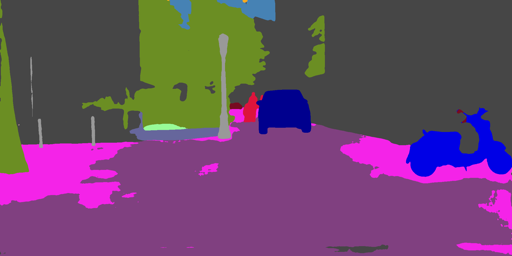}
		\includegraphics[width=\linewidth,trim={0px 60px 0 0px},clip]{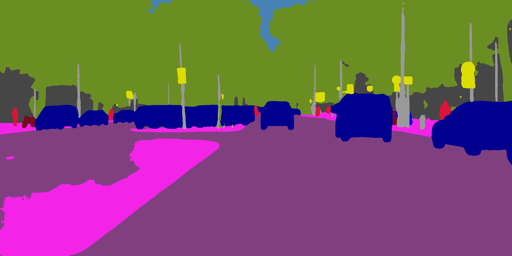}
		\includegraphics[width=\linewidth,trim={0px 60px 0 0px},clip]{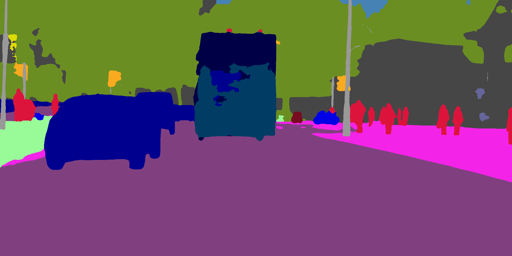}
		\includegraphics[width=\linewidth,trim={0px 60px 0 0px},clip]{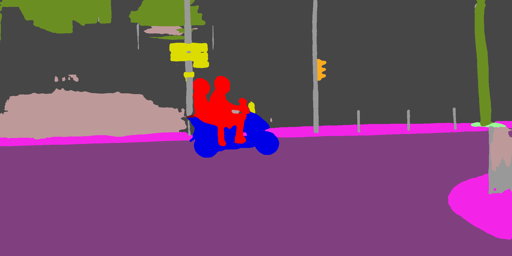}
		\includegraphics[width=\linewidth,trim={0px 60px 0 0px},clip]{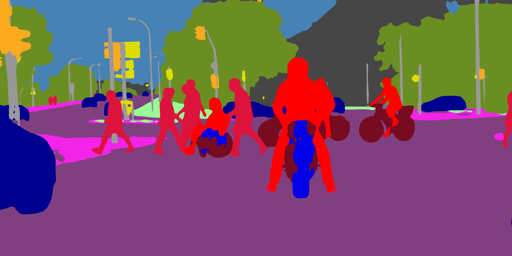}
		\includegraphics[width=\linewidth,trim={0px 60px 0 0px},clip]{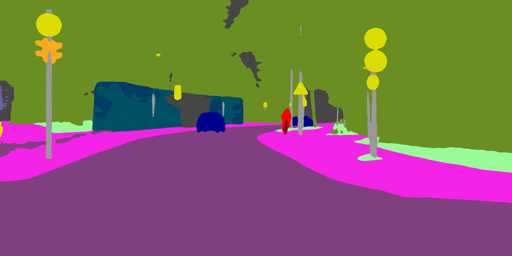}
		\includegraphics[width=\linewidth,trim={0px 60px 0 0px},clip]{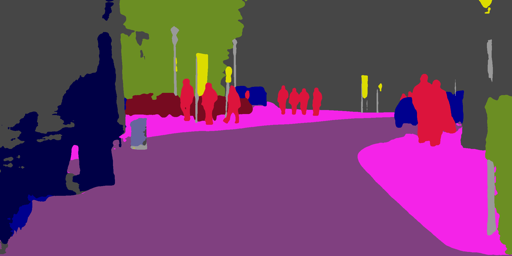}
		\includegraphics[width=\linewidth,trim={0px 60px 0 0px},clip]{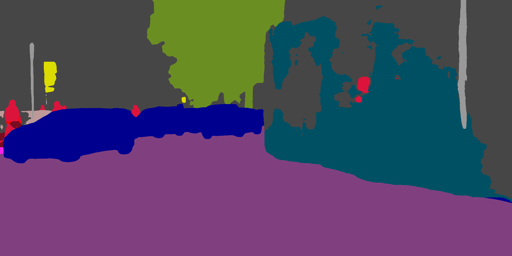}
  \caption{Semantic segmentation}
\end{center}
\end{subfigure}
\begin{subfigure}[t]{0.24\linewidth}
\begin{center}
		\includegraphics[width=\linewidth,trim={0px 60px 0 0px},clip]{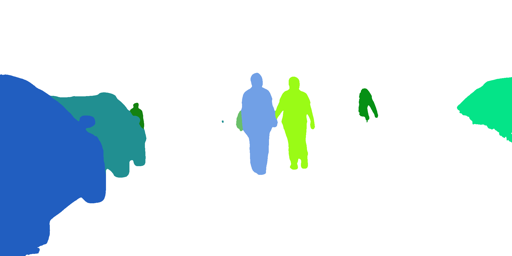}
		\includegraphics[width=\linewidth,trim={0px 60px 0 0px},clip]{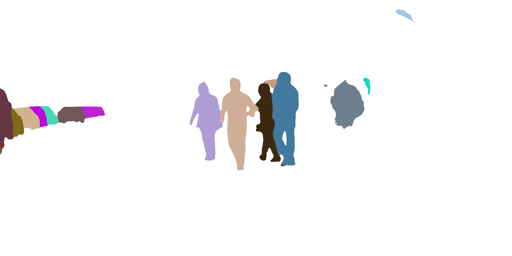}
		\includegraphics[width=\linewidth,trim={0px 60px 0 0px},clip]{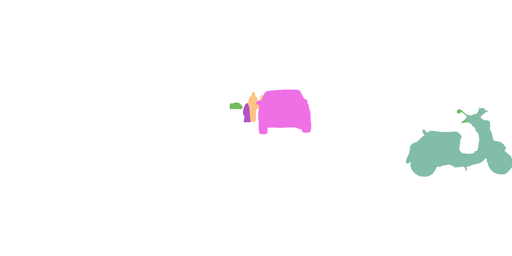}
		\includegraphics[width=\linewidth,trim={0px 60px 0 0px},clip]{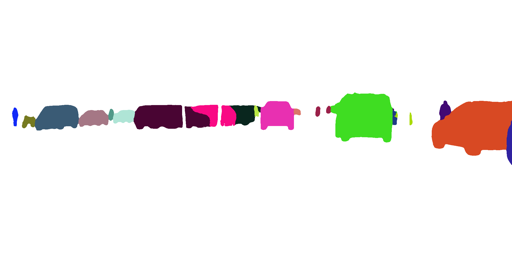}
		\includegraphics[width=\linewidth,trim={0px 60px 0 0px},clip]{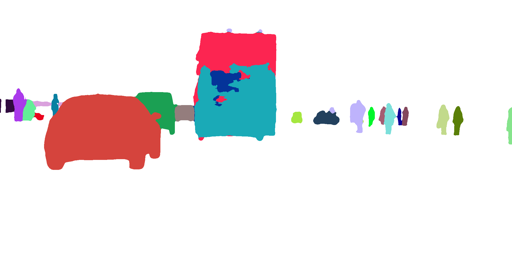}
		\includegraphics[width=\linewidth,trim={0px 60px 0 0px},clip]{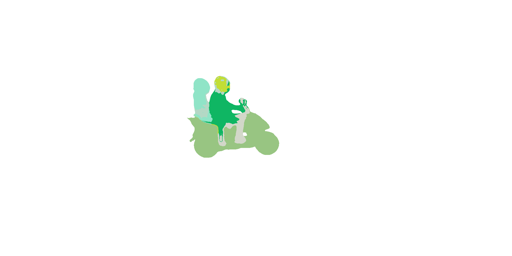}
		\includegraphics[width=\linewidth,trim={0px 60px 0 0px},clip]{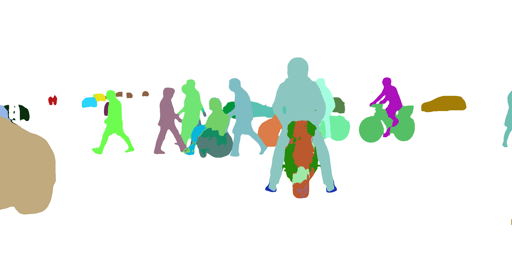}
		\includegraphics[width=\linewidth,trim={0px 60px 0 0px},clip]{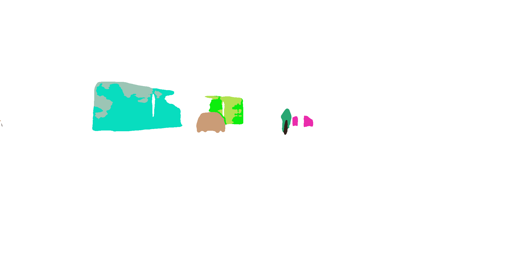}
		\includegraphics[width=\linewidth,trim={0px 60px 0 0px},clip]{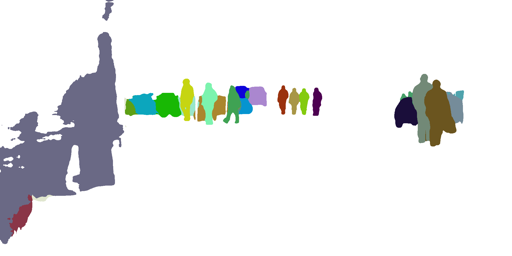}
		\includegraphics[width=\linewidth,trim={0px 60px 0 0px},clip]{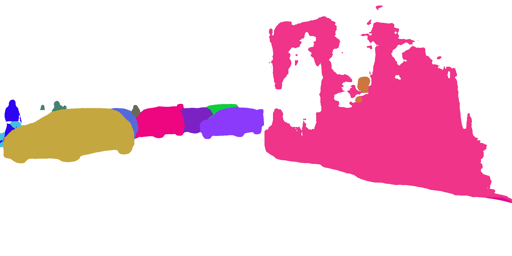}
  \caption{Instance segmentation}
\end{center}
\end{subfigure}
\begin{subfigure}[t]{0.24\linewidth}
\begin{center}
		\includegraphics[width=\linewidth,trim={0px 60px 0 0px},clip]{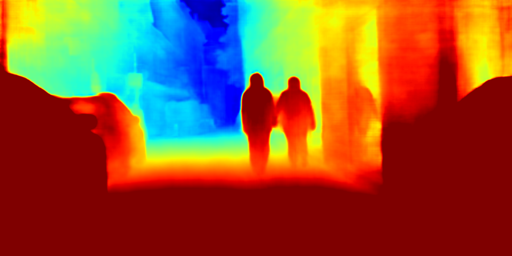}
		\includegraphics[width=\linewidth,trim={0px 60px 0 0px},clip]{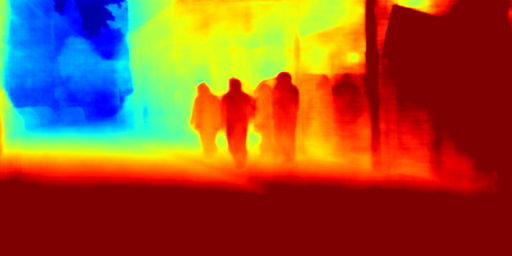}
		\includegraphics[width=\linewidth,trim={0px 60px 0 0px},clip]{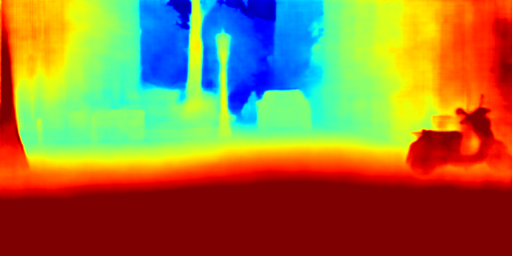}
		\includegraphics[width=\linewidth,trim={0px 60px 0 0px},clip]{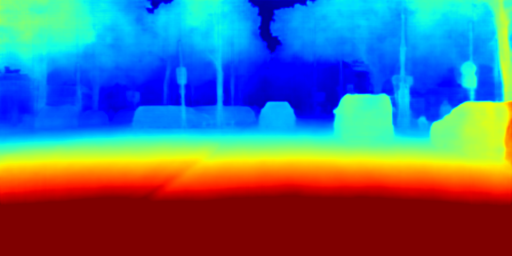}
		\includegraphics[width=\linewidth,trim={0px 60px 0 0px},clip]{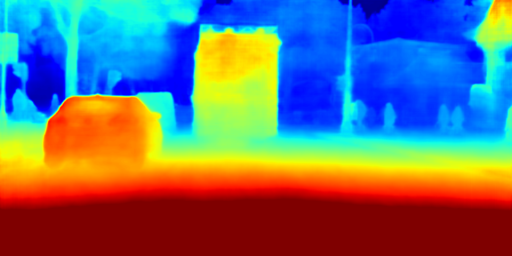}
		\includegraphics[width=\linewidth,trim={0px 60px 0 0px},clip]{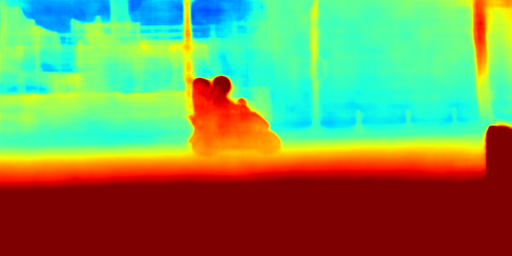}
		\includegraphics[width=\linewidth,trim={0px 60px 0 0px},clip]{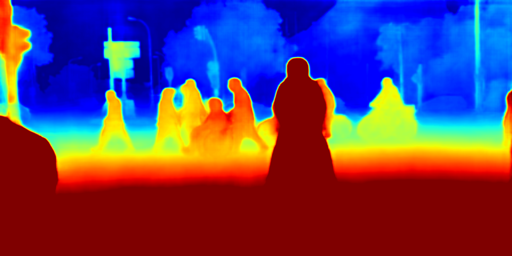}
		\includegraphics[width=\linewidth,trim={0px 60px 0 0px},clip]{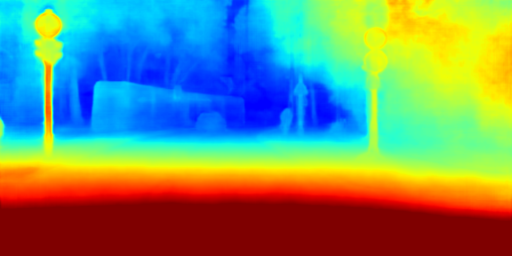}
		\includegraphics[width=\linewidth,trim={0px 60px 0 0px},clip]{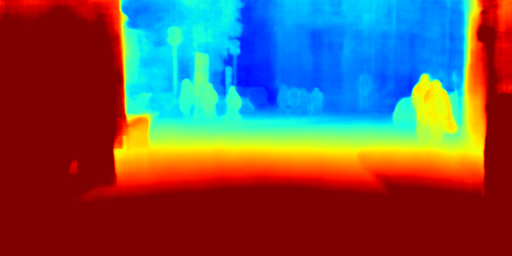}
		\includegraphics[width=\linewidth,trim={0px 60px 0 0px},clip]{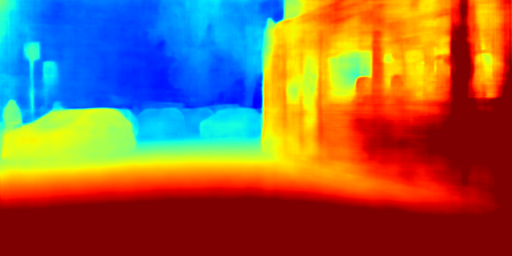}
  \caption{Depth regression}
\end{center}
\end{subfigure}}}
	\caption{\textbf{Example where our model fails on the CityScapes test data.} The first two rows show examples of challenging visual effects such as reflection, which confuse the model. Rows three and four show the model incorrectly distinguishing between road and footpath. This is a common mistake, which we believe is due to a lack of contextual reasoning. Rows five, six and seven demonstrate incorrect classification of a rare class (bus, fence and motorbike, respectively). Finally, the last two rows show failure due to occlusion and where the object is too big for the model's receptive field. Additionally, we observe that failures are highly correlated between the modes, which makes sense as each output is conditioned on the same feature vector. For example, in the second row, the incorrect labelling of the reflection as a person causes the depth estimation to predict human geometry.}
	\label{fig:fail}
\end{figure*}

\end{document}